\documentclass[preprint,12pt]{elsarticle}

\usepackage{amssymb}
\usepackage{makecell}
\usepackage{amsmath}
\usepackage[ruled,vlined]{algorithm2e}
\usepackage{tabularx}
\usepackage{booktabs}
\usepackage{float}
\usepackage{lscape}
\usepackage{adjustbox}
\usepackage{rotating}  
\usepackage{array} 
\usepackage{comment}
\usepackage{amssymb}
\usepackage{tikz}
\usetikzlibrary{positioning}
\usepackage[most]{tcolorbox}
\usepackage{soul}
\usepackage{subcaption}

\newtcolorbox{highLightBox}[1][yellow]{%
  breakable,
  colback=#1!20,
  colframe=#1!20,
  boxrule=0pt,
  arc=0pt,
  left=1pt,right=1pt,top=1pt,bottom=1pt
}

\usepackage{xurl}
\usepackage{afterpage}

\usepackage{caption}
\usepackage{acro}        
\usepackage{pdflscape}
\usepackage{graphicx}
\usepackage{tabularx}
\usepackage{array}
\usepackage{tabularray}
\usepackage{hyperref}

\usepackage{multirow}
\usepackage[table,xcdraw]{xcolor}
\usepackage[normalem]{ulem}
\useunder{\uline}{\ul}{}
\usepackage{lscape}
\usepackage{longtable}
\usepackage{booktabs}
\usepackage{pifont}
\usepackage{adjustbox}

\journal{Nuclear Physics B}

\begin{document}

\begin{frontmatter}




\title{Federated Continual Learning: A Comprehensive Survey on Lifelong and Privacy-Preserving Learning over Distributed and Non-Stationary Data}


\author[label1]{Masoume Gholizade}
\author[label1]{Fabrizio Ruffini}
\author[label1]{Pietro Ducange}
\author[label1]{Francesco Marcelloni}

\affiliation[label1]{organization={Department of Information Engineering, University of Pisa},
             addressline={Largo Lucio Lazzarino 1},
             city={Pisa},
             postcode={56122},
             country={Italy}}


\begin{abstract}

Federated Learning (FL) enables collaborative and privacy-preserving model training across distributed clients, but most existing FL systems implicitly assume data stationarity. In real-world settings—such as healthcare, industrial IoT (IIOT), cybersecurity, and smart cities—data streams are inherently non-stationary, leading classical FL methods to suffer from performance degradation, instability, and catastrophic forgetting.

Continual Learning (CL) addresses learning under evolving data distributions but has been largely studied in centralized settings, overlooking key constraints of federated systems, including privacy, limited communication, and client heterogeneity. Federated Continual Learning (FCL) emerges at the intersection of FL and CL, aiming to support lifelong, adaptive, and privacy-aware learning over distributed and non-stationary data.

This survey provides a comprehensive and systematic overview of FCL. We first present a formal definition of the FCL problem and clarify its distinctive characteristics. We then analyze the limitations of classical FL under non-stationary conditions, highlighting how CL principles support long-term adaptation. To organize the rapidly growing literature, we propose a multi-dimensional taxonomy of FCL approaches. Furthermore, we review representative application domains and data modalities, summarize commonly used evaluation metrics, and discuss experimental perspectives for assessing long-term performance and forgetting. Finally, we highlight key open challenges, including handling extreme heterogeneity under temporal drift, designing scalable and privacy-preserving memory mechanisms, and establishing standardized benchmarks. This survey aims to serve as a reference and a roadmap for advancing FCL toward robust and deployable real-world systems.

\end{abstract}






\end{frontmatter}



\section{Introduction}
\label{sec:introduction}
The proliferation of distributed data across edge devices and organizations has made centralized data collection increasingly impractical due to privacy, regulatory, and communication constraints. Federated Learning (FL) has thus emerged as a transformative paradigm for collaborative model training without requiring the sharing of raw data. By keeping data local to clients and exchanging only model updates, FL enables the construction of global models across distributed data silos while significantly reducing the risks associated with data exposure. This decentralized learning paradigm has been widely adopted in application domains such as mobile and edge computing, healthcare infrastructures, Industrial Internet of Things (IIoT), and smart cities, where sensitive data is continuously generated at scale and strict privacy guarantees are required~\cite{wang2023fedstream}.

Despite these advantages, classical FL algorithms largely assume that client data distributions are stationary, an assumption that rarely holds in real-world environments~\cite{corcuera2022fedxai}. In practical deployments, data streams exhibit continuous distributional shifts due to changes in user behavior, environmental conditions, operational contexts, and the emergence of new classes or tasks. Ignoring such temporal dynamics may result in significant performance degradation, because previously learned knowledge becomes obsolete or is overwritten by subsequent updates~\cite{Kalakoti2025FedXAIIDS}.

The challenge of learning under evolving data distributions has been extensively studied in the context of Continual Learning (CL), which aims to enable models to incrementally acquire new knowledge from streaming data while retaining previously learned information, thereby mitigating catastrophic forgetting. A rich body of literature has explored replay-based strategies, regularization techniques, and architectural adaptations to balance plasticity and stability~\cite{cossu2024continual}. However, most CL approaches are developed under centralized settings, where data is fully accessible and can be stored, replayed, or reorganized without privacy or communication constraints~\cite{li2025kdcl}.

The intersection of FL and CL gives rise to a more challenging paradigm, commonly referred to as Federated Continual Learning (FCL). FCL addresses scenarios in which data is both decentralized across multiple clients and non-stationary over time. A central challenge in FCL is the tension between knowledge retention and strict privacy constraints that limit the storage or sharing of historical data~\cite{chen2025fedmtl}. In such settings, models must cope with catastrophic forgetting, statistical heterogeneity, limited communication budgets, and partial client participation. Consequently, FCL is not merely a straightforward combination of FL and CL, but a distinct paradigm that must balance continual adaptation, collaborative learning, and privacy preservation.

The necessity of FCL becomes particularly evident in high-impact application domains where data evolves continuously and decisions must remain reliable over time. In healthcare, changes in patient populations, clinical protocols, and diagnostic criteria require models that can adapt without forgetting prior medical knowledge. In IIoT and smart manufacturing, sensor data distributions change over time due to equipment aging, maintenance, and evolving operational conditions. In cybersecurity and networking, new attack patterns emerge while historical threats must still be recognized 
~\cite{ZHANG2024108826}. In such domains, data often cannot be centralized, and model updates must be both privacy-aware and robust to temporal changes, making FCL a natural and essential learning paradigm.

Although FCL has recently attracted increasing attention, the existing literature remains fragmented. Many studies focus on specific algorithmic components, particular CL strategies, or narrow experimental settings, making it difficult to derive general design principles or enable systematic comparisons across approaches. Evaluation protocols also vary widely, with no clear consensus on how to jointly assess long-term accuracy, catastrophic forgetting, model stability, communication efficiency, and memory requirements~\cite{Wang2025}.

Prior work has examined related issues such as non-IID data, concept drift, and continual learning in federated environments, highlighting the strong interaction between statistical heterogeneity and temporal dynamics~\cite{criado2022non, yang2024federated}. However, existing surveys typically focus on specific aspects of the problem, such as FL under non-IID conditions or CL in centralized or partially distributed settings, 
without providing a unified and design-oriented perspective on FCL. Table \ref{tab:fcl_comparison} highlights the evolution of surveys in FCL and clearly outlines several limitations in the existing literature. 
Taxonomies are frequently partial or constrained (e.g., fusion-based, task-based, or scenario-specific), and there is a general lack of comprehensive coverage across applications, datasets, and evaluation protocols. In particular, datasets and evaluation strategies are often limited, implicit, or non-standardized, which hinders reproducibility and systematic comparison.

In contrast, this work adopts a unified FCL perspective and introduces a multi-dimensional taxonomy that jointly considers adaptation, aggregation, personalization, and CL paradigms. It also provides a more comprehensive and structured view of applications and datasets, with modality-aware organization and standardized evaluation criteria. 

\begin{table*}[t]
\centering
\caption{Comparison of existing surveys on FCL}
\label{tab:fcl_comparison}
\resizebox{\textwidth}{!}{
\begin{tabular}{l l l l l l l}
\hline
\textbf{Ref. (Year)} & \textbf{Perspective} & \textbf{Taxonomy} & \textbf{Applications} & \textbf{Datasets} & \textbf{Evaluation} & \textbf{Gap} \\
\hline

\cite{criado2022non} (2022) & Problem-focused & $-$ & $-$ & Limited & Empirical & No FCL view \\

\cite{yang2024federated} (2023) & Method-focused & Fusion-based & $-$ & Very limited & Limited & Narrow scope \\

\cite{wang2411federated} (2024) & Domain-focused & Task-based & $\checkmark$ & Implicit & Limited & Domain-specific \\

\cite{hamedi2025fcl} (2025) & Challenge-driven & Partial & $\checkmark$ & Limited & Limited & No unified structure \\

\cite{birashk2025federated} (2025) & Scenario-focused & CL-based & Limited & Scenario-limited & Moderate & Restricted scope \\

\cite{savoia2026federated} (2026) & Application-focused & Application-based & $\checkmark$ & Domain-level & Limited & Application-specific \\

\textbf{This Work} 
& \textbf{Unified FCL perspective} 
& \makecell[l]{\textbf{Adaptation, Aggregation, } \\ \textbf{Personalization, and CL paradigms}} 
& \makecell[l]{\textbf{Comprehensive} \\ \textbf{}} 
& \makecell[l]{\textbf{Structured,} \\ \textbf{modality-aware}} 
& \makecell[l]{\textbf{Standardized} \\ \textbf{and structured}} 
& \textbf{--} \\

\hline
\end{tabular}
}
\end{table*}

More precisely, this survey (i) formalizes the FCL problem, (ii) analyzes the limitations of classical FL in dynamic environments, and (iii) systematically organizes existing approaches through a multi-dimensional taxonomy and a critical evaluation perspective that jointly capture decentralization, continual adaptation, and privacy preservation.

The main contributions of this survey are summarized as follows:
\begin{itemize}
    \item A formal definition of FCL is provided, explicitly distinguishing it from traditional FL and centralized CL settings.
    
    \item A design-oriented, multi-dimensional taxonomy is developed to systematically categorize existing FCL approaches based on adaptation strategies, aggregation mechanisms, personalization techniques, and underlying CL paradigms.
    
    \item An application-driven perspective is presented by reviewing major domains and dataset modalities, highlighting the practical relevance and diversity of FCL scenarios.
    
    \item Experimental protocols and evaluation metrics are critically examined, emphasizing standardized assessment of long-term performance, catastrophic forgetting, communication efficiency, and memory requirements.
    
    \item Key unresolved issues are identified, including heterogeneity under temporal drift, privacy-preserving memory mechanisms, global model stability, and the lack of standardized benchmarks, along with promising directions for future research.
\end{itemize}

\begin{figure}[htbp]
    \centering
    \includegraphics[width=0.85\textwidth]{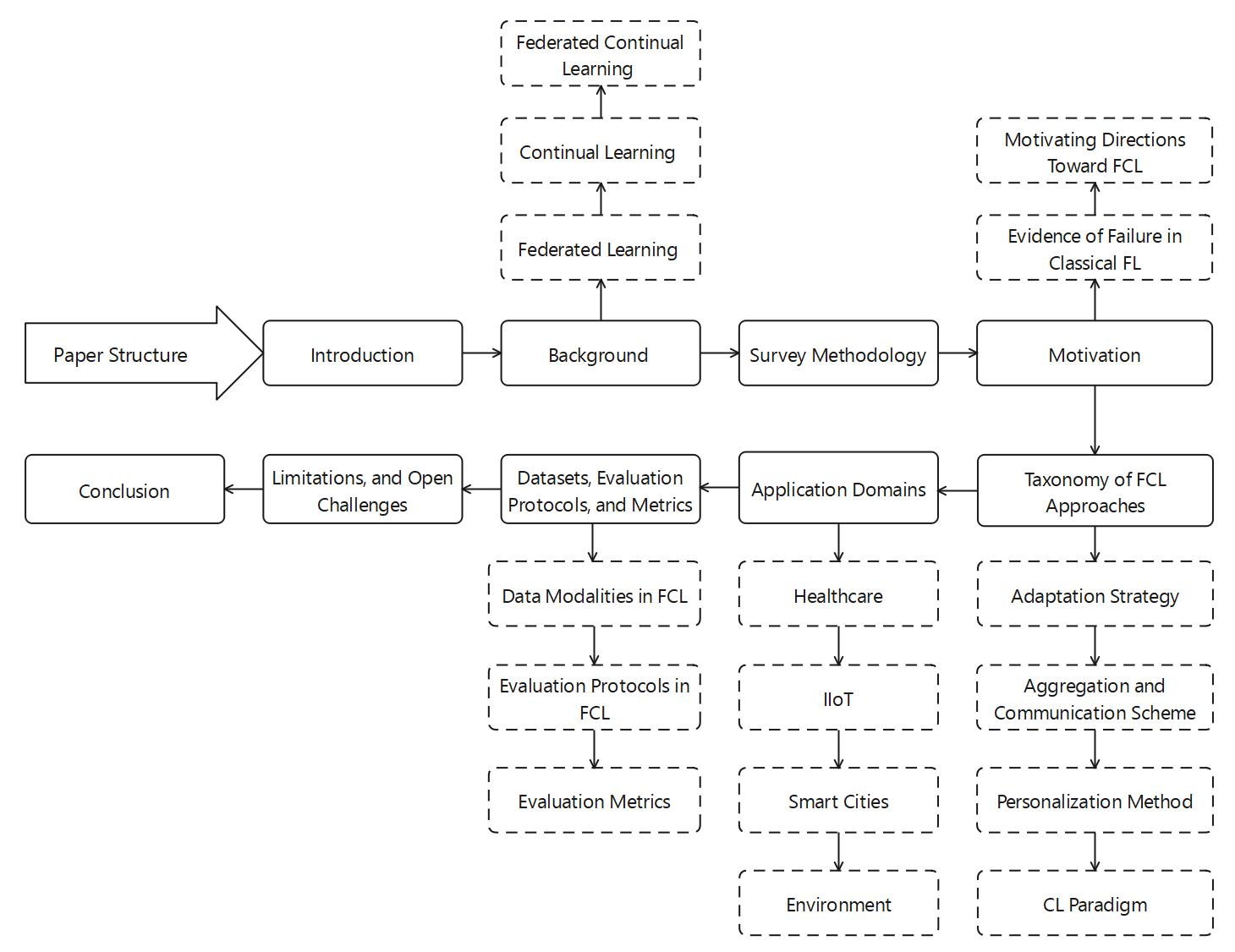}
    \caption{Overall structure of the paper.}
    \label{fig:survey_flowchart}
\end{figure}

The structure of this survey is illustrated in Figure~\ref{fig:survey_flowchart}. Section~\ref{sec:background} reviews the necessary background on FL, CL, and their integration into FCL, followed by the survey methodology in Section~\ref{sec:survey_methodology}. Section~\ref{sec:Motivation-Fed-CL} motivates the need for FCL by analyzing the limitations of classical FL in dynamic environments. Section~\ref{sec:taxonomy} presents a comprehensive taxonomy of existing FCL approaches based on adaptation strategies, aggregation mechanisms, personalization methods, and CL paradigms, while Section~\ref{sec:applic} reviews major application domains and use cases, and Section~\ref{sec:dataset} discusses different dataset modalities in FCL. Section~\ref{sec:limitations-FCL} highlights the limitations, open challenges, and future research directions, and Section~\ref{sec:conclusion} concludes the survey.

\section{Background and Definitions}
\label{sec:background}

This section provides the necessary background for FCL. It first reviews FL from the perspective of non-stationary environments, highlighting its limitations when data distributions evolve over time. It then introduces CL and discusses its limitations in decentralized and privacy-sensitive settings. Finally, it formulates FCL as a unified learning paradigm that integrates decentralization, continual adaptation, and knowledge retention, establishing the conceptual foundation for the remainder of this survey.

\subsection{Federated Learning (FL)}
\label{subsec:fl}

FL is a distributed machine learning paradigm that enables multiple clients to collaboratively train a shared model while keeping training data localized at each participant~\cite{mangye2023hetero}. Instead of collecting raw data in a centralized repository, FL relies on the iterative exchange of model parameters or gradients, which substantially reduces privacy risks and communication overhead. This decentralized training principle has positioned FL as a fundamental solution for privacy-preserving learning in sensitive and large-scale environments such as healthcare, IIoT, mobile edge computing, and smart cities~\cite{Nandi2023DockerFederated}.

The standard FL setting follows a client--server architecture, in which a central server coordinates the learning process and maintains the global model, while a population of distributed clients holds private datasets and performs local training. In each communication round, the server selects a subset of available clients, broadcasts the current global model, and receives locally updated parameters from them. These updates are then aggregated to construct a new global model, which is subsequently redistributed to the clients. This architecture enables scalable collaboration across decentralized data silos while preserving data locality and reducing the direct exposure of sensitive information~\cite{NANDI2022340}.

Formally, let \(K\) denote the total number of clients and let \(\mathcal{D}_k\) represent the local dataset of client \(k\). The global optimization objective in FL is typically formulated as
\begin{equation}
\min_{w} F(w) = \sum_{k=1}^{K} p_k F_k(w),
\end{equation}
where \(w\) denotes the global model parameters, \(F_k(w)\) is the local empirical risk over \(\mathcal{D}_k\), and \(p_k\) is a weighting factor, commonly chosen to be proportional to the cardinality \(|\mathcal{D}_k|\) of \(\mathcal{D}_k\). This formulation highlights that FL optimizes a weighted combination of heterogeneous local objectives. When client data distributions are non-IID, which is the norm rather than the exception in practice, such heterogeneity, combined with additional factors such as privacy-induced noise, can significantly degrade model performance and learning effectiveness~\cite{Cai2024Federated}.

Operationally, FL proceeds through a sequence of iterative communication rounds. In each round, a subset of clients performs several steps of local optimization using their private data, after which the resulting model updates are uploaded to the server and aggregated to update the global model. This protocol significantly reduces communication overhead by limiting participation to only a fraction of clients per round and is particularly well suited to resource-constrained and intermittently connected environments~\cite{Cai2024Federated}.

Federated Averaging (FedAvg) is the most widely used aggregation method due to its simplicity and computational efficiency, computing a weighted average of client models based on the sizes of their local datasets~\cite{Nandi2023DockerFederated}. However, its performance often deteriorates under strong data heterogeneity and partial client participation, which are common in real-world deployments. To address these limitations, various extensions have been proposed, including proximal-based methods, adaptive aggregation strategies, and asynchronous FL protocols~\cite{ID4,ID5}. These challenges become particularly critical in dynamic environments, where data distributions evolve over time, highlighting the need for continual adaptation mechanisms in federated training~\cite{WU2024414}.

FL is inherently characterized by both statistical and system heterogeneity. Statistical heterogeneity arises from the fact that client datasets are typically non-IID, unbalanced, and domain-specific. System heterogeneity results from variations in computational power, memory capacity, energy constraints, and network conditions across clients. These two forms of heterogeneity significantly complicate the optimization process and motivate the design of adaptive aggregation rules, client selection strategies, and resource-aware training protocols~\cite{WU2024414}.

From a privacy perspective, FL reduces the risk of direct data leakage by keeping raw data local. However, model updates can still expose sensitive information through inference and reconstruction attacks. Consequently, FL is often combined with secure aggregation protocols and differential privacy (DP) mechanisms. Recent studies show that integrating DP into continuous or streaming FL can provide formal privacy guarantees under continual updates, albeit at the cost of additional computation and reduced model accuracy~\cite{Jin2021}.

Classical FL methods implicitly assume that client data distributions remain stationary over time. In practice, this assumption rarely holds. Real-world applications such as real-time emotion recognition from wearable devices~\cite{Zhang2023FedSI}, IIOT monitoring systems~\cite{zhou2025cross}, and network traffic analysis platforms~\cite{WU2024414} operate on continuously evolving data streams. Under such non-stationary conditions, naively applying FL can lead to model degradation and catastrophic forgetting, as the global model is repeatedly optimized on shifting and potentially conflicting data distributions.

These limitations have motivated the extension of FL toward FCL, in which continual adaptation mechanisms are embedded into the federated training process. Representative examples include communication-efficient FCL based on synaptic intelligence~\cite{CrossFCL2024}, concept-drift-aware FL frameworks that explicitly detect and adapt to distribution shifts~\cite{shan2025survey}, and parameter-importance-based aggregation strategies that preserve previously learned knowledge across federated rounds~\cite{LiuWeijie2024}. These approaches demonstrate that incorporating CL principles into FL is essential for maintaining long-term model performance in dynamic environments.

\subsection{Continual Learning (CL)}
\label{subsec:cl}

CL is a learning paradigm in which models incrementally acquire and update knowledge from sequential tasks or non-stationary data streams while mitigating \emph{catastrophic forgetting}. In contrast to traditional i.i.d.\ training, CL explicitly seeks to preserve previously learned information (\emph{stability}) while maintaining the ability to integrate new knowledge (\emph{plasticity}). Achieving an appropriate balance between these two objectives is fundamental for building adaptive systems that can operate reliably over long time horizons~\cite{de2021continual}. This stability--plasticity dilemma constitutes the theoretical core of CL and underpins most algorithmic developments in this area.

The need for CL arises from the dynamic nature of real-world environments, where data distributions evolve continuously and retraining models from scratch becomes computationally infeasible or impractical due to memory, privacy, or scalability constraints~\cite{shi2025llmcl}. Modern intelligent systems, ranging from industrial monitoring platforms to large language models, are increasingly deployed in settings where new knowledge must be incorporated without erasing previously acquired capabilities. As a result, CL has become a central building block for long-lived and adaptive machine learning systems.

To systematically study sequential learning, the CL community has proposed several taxonomies that categorize problem settings and methodological strategies. From the problem formulation perspective, CL scenarios are commonly divided into task-incremental, class-incremental, and domain-incremental learning~\cite{de2021continual}. Task-incremental learning assumes that task identity is available during both training and inference: a shared backbone is used to learn common representations across tasks, while task-specific output heads are maintained and selected at inference time based on the known task identity. 
Class-incremental learning is more challenging, as new classes are introduced over time under a shared output space and task identity is unavailable at inference. Domain-incremental learning considers settings in which the label space remains fixed but the input distribution changes due to domain shifts, sensor drift, or environmental variations. These paradigms provide a unified abstraction for capturing different types of non-stationarity in sequential learning environments.

Early research in CL focused on catastrophic interference, where sequential training overwrites previously learned knowledge. This led to several mitigation strategies, broadly categorized into replay-based, regularization-based, representation-based, and architecture-based methods~\cite{de2021continual}. Replay-based approaches reuse past samples to approximate joint training, as in iCaRL~\cite{rebuffi2017icarl} and experience replay methods~\cite{rolnick2019experience}. Regularization-based methods, such as Learning without Forgetting (LwF)~\cite{li2017learning} and Elastic Weight Consolidation (EWC), constrain parameter updates to preserve prior knowledge. Representation-based methods aim to learn robust features that reduce task interference~\cite{krishnan2022self}, while architecture-based approaches allocate task-specific components to isolate knowledge~\cite{IQBAL2025104157,YANG202416}.

Despite their effectiveness in centralized settings, classical CL methods assume unrestricted access to training data and optimization processes, making them unsuitable for large-scale distributed environments where data is decentralized and privacy-sensitive. Moreover, CL does not address coordination among multiple data holders with heterogeneous distributions, resources, and connectivity patterns. While CL enables models to adapt over time, it lacks mechanisms for collaborative learning across independent clients.

These limitations indicate that CL alone is insufficient for scalable and privacy-preserving learning in distributed systems. FL complements CL by enabling collaborative training across decentralized data silos while preserving data locality, thereby addressing the \emph{spatial} dimension of data distribution, whereas CL addresses the \emph{temporal} dimension of data evolution~\cite{zhang2021client}. 
Their integration gives rise to FCL.

\subsection{Federated Continual Learning (FCL)}

FCL unifies the CL and FL paradigms by enabling lifelong model adaptation in decentralized systems while preserving data privacy and mitigating catastrophic forgetting~\cite{zhang2021client}. Hence, FCL aims to learn a global model that can continuously absorb new knowledge from heterogeneous clients while maintaining satisfactory performance on previously learned tasks~\cite{yang2024federated}.

In this work, we adopt a pragmatic and application-oriented view of FCL. 
Beyond classical task-incremental or class-incremental protocols, we also regard FL systems operating on streaming, non-stationary data as instances of FCL when they must preserve historical knowledge while adapting to evolving distributions. 
Although such systems may not explicitly define task boundaries, they exhibit the same fundamental challenges of CL, including catastrophic forgetting, stability–plasticity trade-offs, and long-horizon robustness under decentralized constraints.

\begin{figure}
    \centering
    \includegraphics[width=0.85\linewidth]{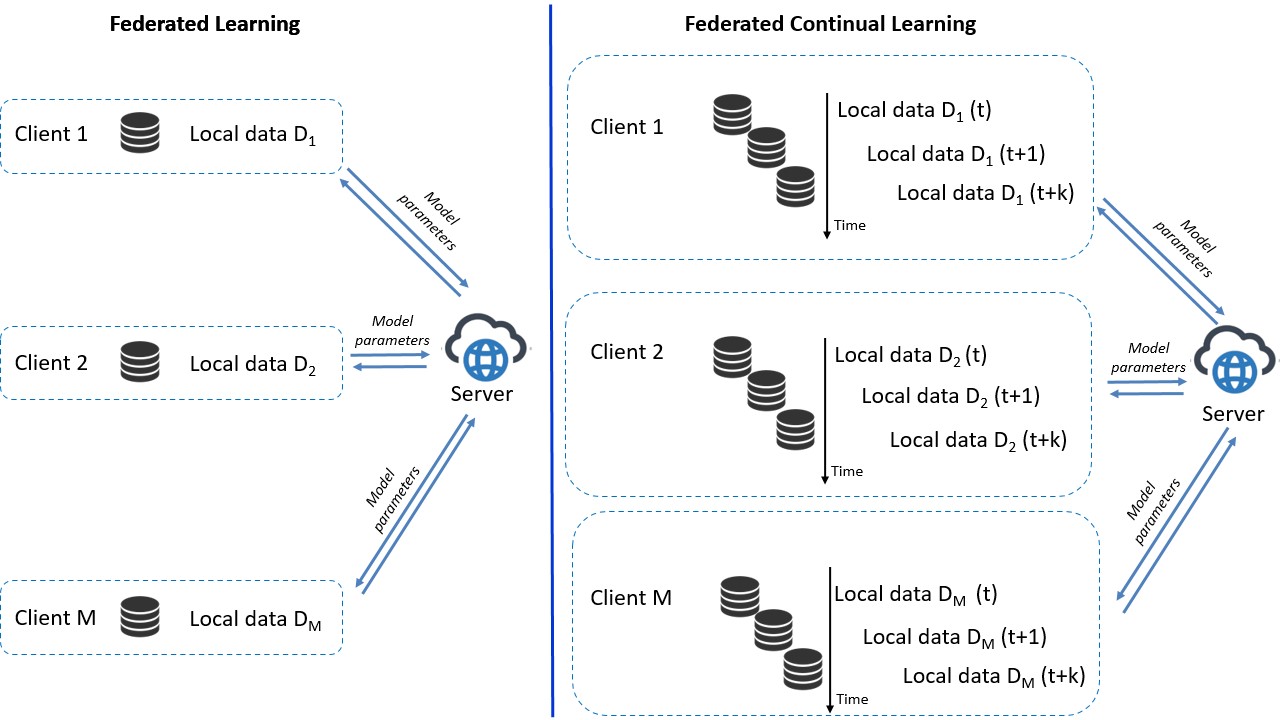}
    \caption{Comparison between FL with stationary datasets (left) and FCL with time-evolving data (right)}
    \label{fig:FL_FCL}
\end{figure}

More formally, FCL is defined as a learning paradigm in which a set of distributed clients ${c_1,\dots,c_N}$ collaboratively train a global model $f_\theta$ under a federated protocol, while each client receives local data streams ${\mathcal{D}_i^1,\mathcal{D}_i^2,\dots}$ that arrive over time and may correspond to new classes, new tasks, or shifted data distributions, where task boundaries can be either explicitly defined or implicitly induced by concept drift, and evolving environments.
Figure \ref{fig:FL_FCL} shows a schematic comparison of the differences between the FL and the FCL paradigms.

The objective of FCL is to optimize $f_\theta$ such that it simultaneously satisfies three requirements: (i) \emph{federation}, meaning that raw data never leave local clients and only model updates or high-level knowledge are exchanged; (ii) \emph{continual adaptation}, meaning that the model can incrementally incorporate new information without retraining from scratch; and (iii) \emph{knowledge retention}, meaning that performance on previously learned tasks is preserved, thus alleviating catastrophic forgetting. This definition distinguishes FCL from classical FL, which focuses on collaborative optimization under static data assumptions, and from centralized CL, which addresses temporal adaptation but ignores privacy, communication constraints, and client heterogeneity. Therefore, FCL represents a fundamentally more challenging problem setting that integrates decentralization, non-stationarity, and memory preservation into a unified framework.

Early efforts toward FCL originated from streaming-aware federated systems that incorporated continual adaptation to cope with evolving data. For example, FedStream demonstrated that dynamic data management and memory-aware updates can mitigate catastrophic forgetting and improve stability in IIOT scenarios~\cite{WU2024414}. Subsequent work introduced more principled solutions that explicitly integrate continual learning techniques into federated settings. Regularization-based frameworks such as FINAL leverage federated Fisher information to preserve long-term knowledge across tasks and communication rounds~\cite{Zhao2024FINAL}, while representation learning approaches, exemplified by EvoFedIDS, employ contrastive learning and memory-aware updates to maintain discriminative features under evolving data distributions~\cite{ZHANG2024108826}.

More recently, generative replay strategies have been explored to enable privacy-preserving memory without storing sensitive historical data. For example, diffusion-based frameworks can synthesize representative samples of past tasks, thus mitigating catastrophic forgetting while maintaining compliance with data privacy constraints~\cite{Sun2025Federated}. Overall, these studies illustrate a clear progression from heuristic adaptations of FL toward theoretically grounded and memory-aware learning frameworks, establishing FCL as a key paradigm for adaptive learning in non-stationary and decentralized environments.

\section{Survey Methodology}
\label{sec:survey_methodology}

\begin{figure}[htbp]
\centering
\includegraphics[width=0.95\textwidth]{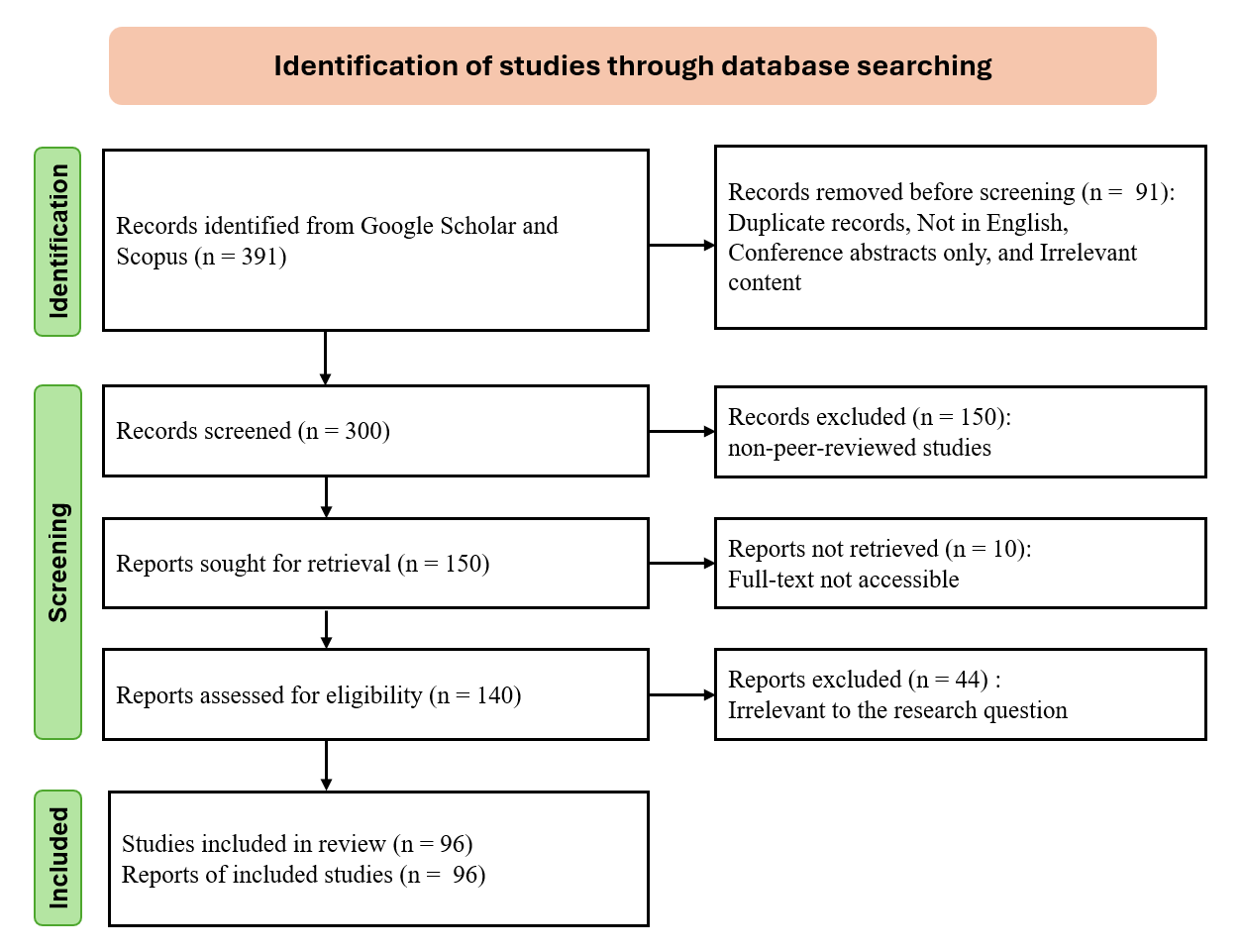}
\caption{PRISMA flow diagram of the literature selection process.}
\label{fig:prisma}
\end{figure}

This survey is conducted as a systematic literature review following a PRISMA-inspired methodology~\cite{PRISMA} to ensure transparency, reproducibility, and methodological rigor. The overall workflow consists of three main stages: (i) keyword definition, (ii) source selection, and (iii) multi-stage screening and eligibility assessment. The study selection process is illustrated in Fig.~\ref{fig:prisma}.

\subsection{Keyword Definition}
A structured search strategy was developed to comprehensively cover the FCL domain. The queries combined ``federated learning'' with key continual learning concepts, including ``continual learning'', ``incremental learning'', and ``catastrophic forgetting''.

To address terminological variations, additional terms such as ``lifelong learning'' and ``continuous learning'' were incorporated. All keywords were combined using Boolean operators (AND, OR) and applied to titles, abstracts, and keywords across databases.

\subsection{Source Selection}
The literature search was conducted using Google Scholar to ensure broad coverage and maximize recall. Multiple keyword combinations were applied, resulting in 314 records prior to screening.
To complement this search with high-quality peer-reviewed papers, Scopus was used as a curated database with the same queries, yielding 77 records before deduplication.

\begin{figure}[htbp]
\centering
\includegraphics[width=0.95\textwidth]{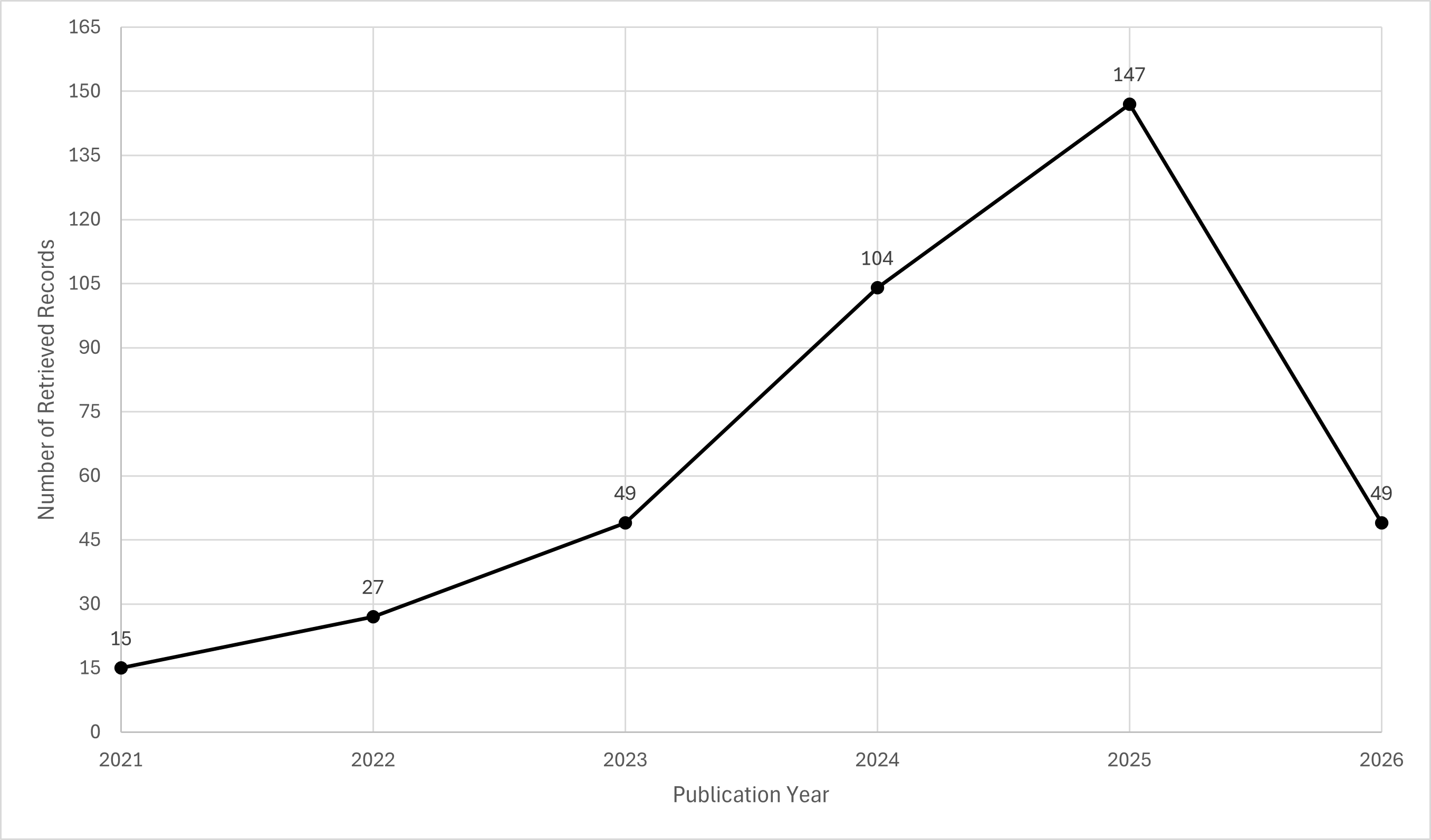}
\caption{Temporal distribution of the initial search results retrieved from Scopus and Google Scholar.}
\label{fig:year_distribution}
\end{figure}

All searches were performed up to March 2026. Most relevant papers were published after 2021, indicating that FCL is an emerging research area. The temporal distribution of the relevant papers is presented in Fig.~\ref{fig:year_distribution}.

\subsection{Multi-stage screening and eligibility assessment}
After removing duplicate and irrelevant entries, a multi-stage screening process was conducted, including title/abstract filtering and full-text assessment.
Studies were included if they addressed continual or incremental learning in federated settings, proposed methodological contributions, and provided experimental validation. Studies focusing on a single paradigm, non-peer-reviewed works, and incomplete or duplicate publications were excluded.
Following the screening and eligibility phases, 96 studies were selected for the final analysis. Backward snowballing was also performed to ensure completeness of the review.

A qualitative assessment was conducted to verify the methodological soundness of the selected studies. 

\section{Motivation for Federated Continual Learning}
\label{sec:Motivation-Fed-CL}

This section provides a problem-driven motivation for FCL by systematically analyzing the fundamental limitations of classical FL in non-stationary and dynamic environments. It reviews empirical evidence and recent studies showing that standard FL algorithms suffer from instability, catastrophic forgetting, and degraded convergence when data distributions evolve over time, clients participate asynchronously, and resource and privacy constraints are imposed. These observations collectively demonstrate that classical FL lacks the structural mechanisms required for continual adaptation, long-term memory retention, and robustness under temporal distribution shifts, thereby motivating the emergence of FCL as a necessary learning paradigm.

\subsection{Evidence of Failure in Traditional FL}

Traditional FL typically assumes (i) relatively stable local data distributions and (ii) periodic global aggregation under well-behaved client participation. In contrast, real-world deployments feature evolving local datasets, intermittent client availability, and temporal distribution drift. Early empirical evidence of these limitations is provided by the FCL-BL framework~\cite{Le2021}, which shows that successive global updates can diverge substantially when clients join or leave unpredictably, causing instability and degradation of previously learned knowledge.
The ECFL framework~\cite{casado2023ensemble}  further reinforces this motivation by showing that even ensemble-based aggregation can suffer from severe inter-round instability under distribution shift. 

Traditional FL algorithms such as FedAvg are fundamentally limited in environments characterized by
drift, continual updates, and heterogeneous client behavior. As highlighted in FCL-BL~\cite{Le2021},
synchronous FL suffers from stragglers, while asynchronous updates introduce stale gradients that
degrade global convergence. More critically, sequential optimization in standard FL tends to
overwrite previously learned representations, resulting in catastrophic forgetting. ECFL~\cite{casado2023ensemble}
further shows that even decision-level fusion becomes unstable when data distributions shift, as its
confidence-weighted voting cannot reliably compensate for drift. These observations indicate that
traditional FL lacks the structural components required for continual knowledge preservation,
drift awareness, and adaptive learning---limitations that directly motivate the development of FCL.

\subsection{Motivating Directions Toward FCL}
Building on the observed limitations of classical FL, several research directions have emerged to enable continual adaptation in federated settings.

A major direction in FCL aims to mitigate the combined effects of non-IID data and client drift through parameter-space constraints. Regularization-based approaches attempt to preserve previously learned knowledge by limiting excessive updates to important parameters, thereby stabilizing training under heterogeneous and evolving data distributions. These strategies are largely inspired by CL methods that address the stability–plasticity dilemma~\cite{Zhang2023FedSI}.

Complementing parameter-space approaches, another direction focuses on representation-level adaptation. These methods summarize evolving data into compact representations that enable drift detection and continual adaptation while avoiding the need to share raw data across clients. Such strategies are particularly suitable for privacy-sensitive and communication-constrained environments~\cite{Mawuli2023}.

A further line of research addresses scenarios with distribution drift and limited label availability, motivating semi-supervised and weakly-supervised CL strategies in federated environments~\cite{Cobbinah2023}.

In parallel, importance-aware approaches estimate which parameters encode prior knowledge and constrain their updates, thereby mitigating catastrophic forgetting while enabling adaptation to evolving data streams~\cite{Zhao2024FINAL}.
Recent research also explores importance-aware aggregation and memory-based mechanisms to enhance long-term stability in FCL. By reducing updates to parameters that encode prior knowledge and maintaining compact representations of past tasks, these approaches aim to mitigate catastrophic forgetting while preserving privacy and communication efficiency under severe distribution shifts~\cite{Yu2025,WANG2023551}.

A major driver behind the emergence of FCL lies in the fundamental limitations of classical FL in dynamic and streaming environments. 
Streaming clients often operate with limited buffers and rapidly evolving data distributions, leading to mismatches between observed samples and the underlying long-term distribution. As a result, models trained on such transient data exhibit biased updates, unstable convergence, and progressive degradation of previously learned knowledge.

Furthermore, asynchronous data arrivals, fluctuating computation resources, and intermittent connectivity produce inconsistent update trajectories across clients, amplifying drift and destabilizing global optimization. These challenges indicate that classical FL is fundamentally ill-suited for streaming and temporally evolving environments, motivating the need for CL mechanisms capable of online adaptation, drift awareness, and stable long-horizon learning~\cite{Jin2021,LiuWeijie2024,WangHeqiang2024,WangNaiyu2024}.

In addition to non-stationarity and streaming dynamics, FCL is shaped by core federated constraints such as privacy requirements, limited on-device resources, fairness under uneven participation, and uncertainty arising from evolving and partially observed data. Over long training horizons, these factors can amplify forgetting, destabilize aggregation, and bias the global model toward dominant clients or transient patterns. Consequently, effective FCL systems must balance continual adaptation with robustness, efficiency, and fairness while respecting strict privacy constraints~\cite{BACCARELLI2022376,Zeng2023Hfedms,XIONG2026129882}.

Collectively, these findings indicate that FCL must address temporal privacy risks, severe resource limitations, fairness under uneven participation, and heightened uncertainty in evolving data streams. These factors make CL in federated settings substantially more challenging than in centralized environments and highlight the need for mechanisms that jointly balance adaptation and long-term stability.

\section{Taxonomy of FCL Approaches}
\label{sec:taxonomy}

\begin{figure}[htbp]
    \centering
     \includegraphics[height=0.4\textheight]{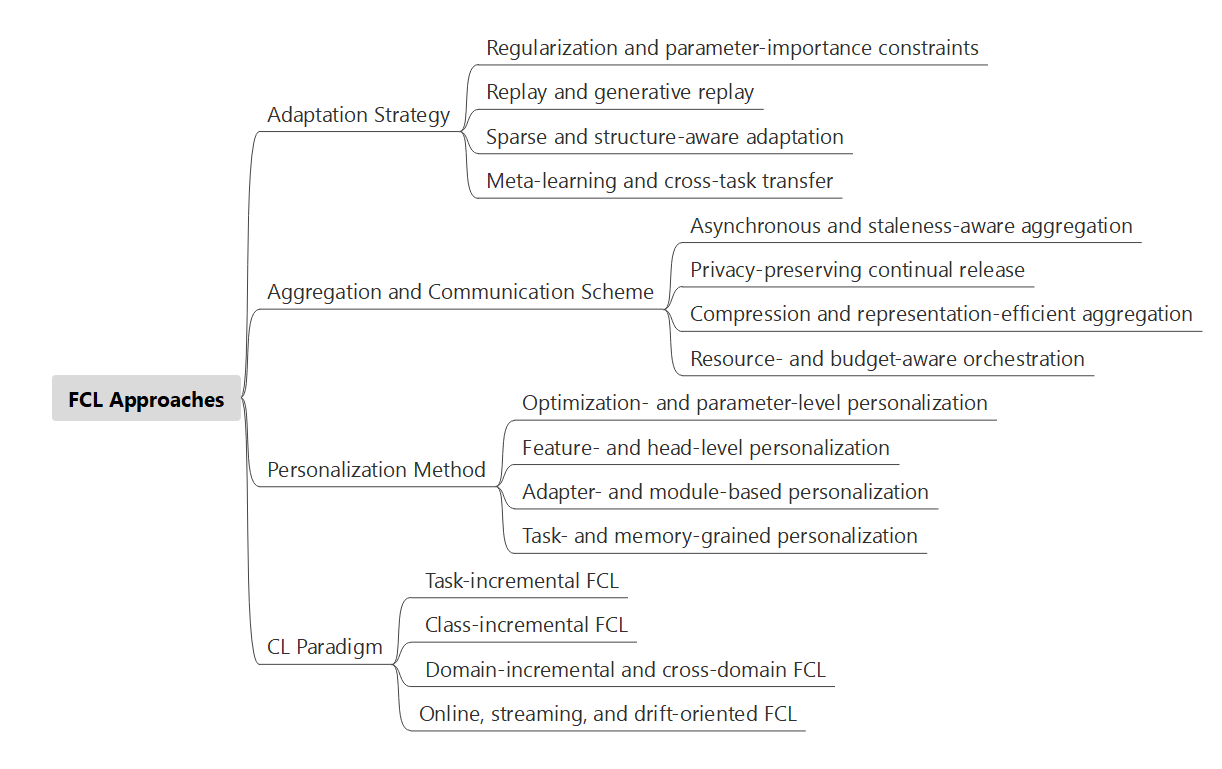}
    \caption{Overall taxonomy of FCL approaches.}
    \label{fig:Picture1}
\end{figure}

The rapidly growing literature on FCL spans diverse algorithmic designs, system assumptions, and application scenarios. 
As illustrated in Fig.~\ref{fig:Picture1}, our taxonomy organizes existing methods along four complementary axes:
(1) adaptation strategy,
(2) aggregation and communication scheme,
(3) personalization mechanism, and
(4) CL paradigm. 
This taxonomy highlights how different techniques address key challenges of FCL, including catastrophic forgetting, statistical and system heterogeneity, communication constraints, privacy requirements, and evolving data distributions.

Because these challenges are often addressed simultaneously in practical systems, many methods naturally span multiple dimensions of this taxonomy.

To make the framework explicitly problem-driven, each axis corresponds to a major source of difficulty in FCL. Adaptation strategies primarily regulate the stability--plasticity trade-off and mitigate forgetting at the client level. Aggregation and Communication schemes address system heterogeneity, limited bandwidth, and intermittent participation, often incorporating asynchronous or staleness-aware updates. Personalization mechanisms handle persistent statistical heterogeneity by isolating client-specific components to prevent negative transfer. Finally, CL paradigms characterize the type of non-stationarity itself (e.g., task-, class-, domain-incremental, or streaming drift), which strongly influences the suitability of different methods.

More concretely, adaptation strategies determine how local models incorporate new knowledge while preserving prior information; aggregation schemes govern how updates are exchanged and combined under resource and privacy constraints; personalization mechanisms maintain client-specific parameters or modules that are not globally aggregated; and CL paradigms characterize the problem setting rather than the learning mechanism. 

Importantly, we treat these four axes as complementary rather than mutually exclusive. In particular, Adaptation strategy, Aggregation and Communication scheme, and Personalization method represent algorithmic design dimensions, whereas the CL paradigm axis describes the underlying non-stationary learning scenario.
Because practical FCL systems typically combine multiple elements, 
each approach is assigned one primary label along each axis, rather than being forced into a single global category.
To ensure consistency between the taxonomy, tables, and textual descriptions, the \emph{primary label} is assigned according to the mechanism that plays the most central functional role in the method.
Operationally, the primary label is determined based on: (i) the core novelty emphasized in the original work, (ii) the main source of performance improvement, and (iii) the component whose removal would most significantly degrade performance or alter the method’s formulation.
Additional mechanisms are treated as secondary characteristics and are discussed when relevant, but they do not determine the primary categorization.

This axis-wise labeling strategy 
provides a transparent and reproducible criterion for categorizing methods that integrate multiple mechanisms.
For example, CFedSI integrates regularization-based adaptation with communication-efficient compression; however, it is categorized under the Adaptation Strategy axis because regularization constitutes its core contribution, while compression is treated as a secondary characteristic under the Aggregation and Communication axis.
This rule ensures consistency between the taxonomy, tables, and textual descriptions, while preserving the inherently multi-dimensional nature of FCL methods.
To further enhance the design-oriented perspective, each category is accompanied by a discussion of its typical application scenarios, core assumptions, and main limitations, providing a systematic understanding of when and why each approach is effective.

\subsection{Adaptation Strategy}

Adaptation strategies determine how local models acquire new knowledge while preserving previously learned information under sequential tasks and heterogeneous clients. 
Operationally, under the Adaptation Strategy axis, an approach is categorized according to the mechanism that primarily controls how new knowledge is incorporated while previously learned information is preserved at the client level.

Unlike centralized CL, the federated setting introduces additional constraints such as limited communication, privacy requirements, and partial participation. Consequently, FCL adaptation strategies must simultaneously balance stability, plasticity, and system efficiency. In the literature, four dominant strategies have emerged: regularization and parameter-importance constraints, replay and generative replay, sparse and structure-aware adaptation, and meta-learning and cross-task transfer.

\paragraph{(1) Regularization and parameter-importance constraints}

Regularization-based methods are among the earliest and most influential approaches for stabilizing local updates in FCL. They formulate CL as a constrained optimization problem, where parameter updates are guided by importance estimates to preserve prior knowledge and mitigate catastrophic forgetting in heterogeneous federated settings.

Early work such as the FCL-BL framework proposed in~\cite{Le2021} uses an analytically tractable pseudo-inverse update rule that evolves the broad learning weights incrementally, avoiding full retraining and providing transparent parameter control in asynchronous FL settings. 

Building on more explicit measurements of parameter relevance, the FedSI and CFedSI frameworks proposed by Zhang et al.~\cite{Zhang2023FedSI} employ Synaptic Intelligence (SI) to quantify the contribution of each parameter along the training trajectory. 
A key component of these methods is the estimation of parameter importance, which determines how strongly each parameter should be preserved during subsequent updates.

Let $g_i^{(t)} = \frac{\partial \mathcal{L}^{(t)}}{\partial \theta_i}$ denote the gradient of the loss function with respect to parameter $\theta_i$ at training step $t$, and let $\Delta \theta_i^{(t)}$ represent the corresponding update of that parameter at the same step.

The contribution of each parameter to the change in the loss can be approximated as $g_i^{(t)} \Delta \theta_i^{(t)}$. Accumulating this contribution over the entire training trajectory, the cumulative contribution of parameter $\theta_i$ is computed as
\begin{equation}
w_i = \sum_t g_i^{(t)} \Delta\theta_i^{(t)}.
\end{equation}

The importance $\Omega_i$ of parameter $\theta_i$ is then defined as
\begin{equation}
\Omega_i = \max \left( \frac{w_i}{(\Delta \theta_i)^2 + \epsilon},\ 0 \right),
\end{equation}
where $\Delta \theta_i = \sum_t \Delta \theta_i^{(t)}$ denotes the total change of parameter $\theta_i$ over the entire training process, and $\epsilon$ is a small damping constant introduced to avoid numerical instability.
Intuitively, $\Omega_i$ captures how much each parameter contributes to the reduction of the loss during training.

Beyond SI-based approaches, Fisher-information–based measures provide an alternative way of estimating parameter importance and have shown strong performance in highly non-IID settings.
The FINAL framework proposed by Zhao et al.~\cite{Zhao2024FINAL} aggregates local Fisher matrices to construct a global importance representation, which is incorporated into the federated update rule to preserve critical knowledge over time. Complementarily, the PI-Fed method introduced by Yu et al.~\cite{Yu2025} aggregates parameter-level relevance across clients and applies gradient soft-masking to protect sensitive weights. The Fisher-based approaches are attractive because they require minimal architectural modification while providing principled mechanisms for controlling parameter drift, often accompanied by theoretical guarantees under standard assumptions.

The methods in this category are particularly suitable for scenarios with strict communication and memory constraints, where storing or generating past data is infeasible, as it relies on the assumption that parameter-importance estimates remain relatively stable across tasks. However, under rapid distribution shifts or highly heterogeneous client updates, these estimates may become unreliable, leading to degraded long-term retention.

\paragraph{(2) Replay and generative replay}

By reconstructing or approximating past data distributions, replay-based methods shift CL from a parameter-constrained problem to a data-centric retention paradigm.
While regularization-based methods constrain parameter updates directly, replay-based approaches instead preserve previously learned knowledge through data reconstruction or rehearsal mechanisms. This family of methods relies on replay, generative replay, or memory-based rehearsal to mitigate forgetting across sequential tasks.

Compared with centralized CL, where replay typically involves storing historical samples, FCL must rely on privacy-preserving surrogates of past data. Direct buffering of raw samples is often infeasible due to privacy regulations and communication constraints. Consequently, recent approaches employ generative models, micro-clusters, semantic summaries, or other compact representations to capture historical knowledge without exposing sensitive data.

Under federated privacy constraints, methods such as CDA-FedAvg~\cite{criado2022non} combine lightweight rehearsal with drift detection. A CUSUM-inspired mechanism detects distribution shifts and activates replay only when necessary, enabling selective reuse of past knowledge.

Generative replay provides a more flexible alternative by synthesizing data instead of storing it. For example, FCL4DD~\cite{Sun2025Federated} employs a weakly supervised diffusion model to generate class-consistent pseudo-samples, compensating for missing historical data in decentralized medical settings. In this paradigm, past information is preserved through synthetic sample generation via stochastic processes such as diffusion models.
Diffusion-based replay follows an iterative reverse denoising process. Starting from a noisy sample $\mathbf{x}_t$, the reconstruction of $\mathbf{x}_{t-1}$ is modeled as:
\begin{equation}
p_\theta(\mathbf{x}_{t-1} \mid \mathbf{x}_t, c, D)
= \mathcal{N}\big(
\mu_\theta(\mathbf{x}_t, t) + \lambda_t f(c, D),
\Sigma_\theta(\mathbf{x}_t, t)
\big),
\end{equation}
where $\mu_\theta(\mathbf{x}_t, t)$ and $\Sigma_\theta(\mathbf{x}_t, t)$ denote the predicted mean and variance, $f(c, D)$ encodes class and diversity information, and $\lambda_t$ controls its influence. This formulation enables the generation of task-consistent samples while preserving diagnostic knowledge without requiring access to raw data.

In industrial monitoring, Li et al.~\cite{LI2024111491} introduce TRINA, which employs self-challenge rehearsal and gradient balancing to correct spatiotemporal drifts across edge devices. Meanwhile, the FedMuL framework proposed by Lamptey et al.~\cite{Lamptey2025} employs micro-clusters as compact memory units to summarize evolving multilabel distributions, enabling scalable replay without raw data retention. Across these works, replay mechanisms reconstruct or approximate past distributions, thereby supporting more robust knowledge retention under severe concept drift.

Recent advances further extend generative replay by integrating more expressive generative models and knowledge transfer mechanisms. For example, diffusion-based frameworks such as FedCapD~\cite{Iqbal2026} leverage generative replay with structured latent representations to improve robustness under non-IID and privacy-constrained settings. In parallel, multi-teacher distillation approaches such as FedMTL~\cite{Chen2026} reformulate replay as a knowledge transfer process, where historical information is preserved through teacher ensembles rather than explicit data reconstruction. Taken together, these developments indicate a shift from purely data-centric replay toward hybrid paradigms that combine generative modeling and knowledge distillation for more scalable and flexible continual adaptation.

The methods in this category are particularly advantageous in environments characterized by severe concept drift and long task horizons, where reconstructing past data distributions becomes essential. However, their effectiveness depends critically on the fidelity of the generated or stored surrogates. Moreover, they introduce additional memory, computational, and communication overhead, especially in resource-constrained federated settings.

\paragraph{(3) Sparse and structure-aware adaptation}

A third adaptation direction focuses on sparse and structure-aware learning, where continual adaptation is reformulated as a selective parameter update problem in which only task-relevant subnetworks or modules are activated, thereby reducing interference between tasks.

In~\cite{Liu2025Spars}  the authors introduce Sparse-FCL, in which dynamic sparse training identifies critical neurons through progressive selection, preserving stable subnetworks while enabling plastic adaptation in resource-limited mobile edge environments. This neuron-level sparsity significantly reduces communication costs and limits interference between tasks.

Structural modularization provides another compelling solution. In~\cite{YANG202416} the authors propose an Adapter--Retainer decomposition, which explicitly divides the model into a trainable module for new knowledge and a frozen module that retains previous representations. This decomposition naturally aligns with device heterogeneity, reduces redundant training, and mitigates the negative influence of old knowledge during early adaptation. 
The Adapter--Retainer is a representative example of the multi-axes labeling rule. Indeed,  it is also categorized as an explicit personalization method because the Adapter remains client-specific and is never globally aggregated, whereas the Retainer serves as a shared knowledge backbone. In this sense, the method spans multiple axes, but receives a distinct primary label within each axis according to the functional role played by each component.
Beyond mitigating catastrophic forgetting, this design also improves system efficiency by restricting updates to compact, task-relevant components, thereby reducing communication overhead, computational cost, and energy consumption.

The methods in this category are particularly effective in resource-constrained and large-scale deployments, where communication efficiency and modular updates are critical, because isolating task-relevant subnetworks reduces interference and unnecessary updates. This benefit, however, relies on the assumption that such subnetworks can be clearly identified and maintained across tasks; otherwise, excessive sparsity or uncontrolled module growth may reduce knowledge sharing, limit forward transfer, and lead to capacity fragmentation as the number of tasks increases.

\paragraph{(4) Meta-learning and cross-task transfer}

A growing body of work leverages meta-learning and cross-task knowledge transfer to accelerate adaptation in federated continual settings, by reframing CL as a learn-to-learn problem in which models acquire transferable initialization or adaptation rules that enable efficient generalization to new tasks.
Unlike regularization- and replay-based approaches, which primarily focus on preserving past knowledge and preventing catastrophic forgetting (backward stability), meta-learning methods explicitly optimize for forward transfer and rapid task adaptation. They aim to learn task-invariant representations or adaptation rules that enable clients to generalize efficiently to unseen tasks with minimal local updates. In this sense, meta-learning introduces a proactive adaptation paradigm in FCL, shifting the objective from merely retaining historical knowledge to accelerating future learning under evolving and heterogeneous data distributions.

The MeCo framework proposed in~\cite{Chen2025know} constructs a task-invariant meta-knowledge pool at the server and learns a retrieval mechanism based on task embeddings, enabling clients to initialize from relevant prior knowledge and achieve faster convergence under evolving industrial tasks. In personalized biometrics,~\cite{Li2023per} proposes cTD-$\alpha$MAML, which combines meta-learning with exemplar-guided distillation to learn rapid adaptation rules while enforcing identity-preserving consistency across tasks. For streaming physiological signals,~\cite{SUN2023586} introduces MetaCL, in which server-side meta-training yields a global feature extractor, and clients maintain lightweight classifiers for continual personalization in a blockchain-secured environment. These meta-learning approaches explicitly model generalization across tasks, offering strong forward-transfer capabilities and significantly reducing the need for extensive local retraining in federated continual environments.

The methods in this category are particularly effective in environments with recurring structure across tasks, where rapid adaptation is required, as it relies on the existence of transferable meta-knowledge across clients and tasks. However, in highly heterogeneous settings, it may suffer from negative transfer and instability, often requiring additional mechanisms such as task-aware routing or personalization.

\subsection{Aggregation and Communication Scheme} 

Aggregation and Communication schemes determine how and when information is exchanged between clients and the server under bandwidth, availability, and privacy constraints. In FCL, these mechanisms must additionally cope with system heterogeneity, partial participation, and evolving data streams. Existing approaches can be broadly categorized into asynchronous aggregation, privacy-preserving continual release, communication-efficient compression, and resource-aware orchestration.
Operationally, under the Aggregation and Communication axis, a method is categorized according to the mechanism that primarily governs when, what, and how information is exchanged and combined across clients and the server.

\paragraph{(1) Asynchronous and staleness-aware aggregation}

In heterogeneous and streaming environments, synchronous aggregation often leads to idle waiting, excessive communication overhead, and degraded system utilization due to stragglers or intermittent participation. Asynchronous aggregation relaxes strict synchronization by allowing clients to update the global model at different times, thereby improving communication efficiency and system utilization under heterogeneous and dynamic conditions.

Le et al.~\cite{Le2021} address this issue through a batch-asynchronous aggregation mechanism in the FCL-BL framework, where the server performs global aggregation only after receiving a predefined number of local updates. By decoupling aggregation from fixed communication rounds, this strategy reduces synchronization delays and allows faster clients to proceed without being blocked by slower ones, making it particularly suitable for CL scenarios with fluctuating client availability.

Building on this direction, Wang et al.~\cite{WangNaiyu2024} propose the FedStream framework for streaming FL, whose asynchronous variant employs an optimization scheme explicitly designed for evolving data streams. In this setting, each client update is assigned an importance weight based on its computational effort, data volume, and staleness. Specifically, the importance coefficient
\begin{equation}
    Q(k) = \frac{n_t(k) + \rho(k)}{\sum_{j \in \mathrm{Eff}} \big(n_t(j) + \rho(j)\big)}
\end{equation}
depends on the local sample size $n_t(k)$ and the number of local iterations $\rho(k)$, while staleness is incorporated via an age factor $\beta(k)$, which influences a separate scaling function $s(k)$ that down-weights outdated updates. This weighting scheme prioritizes contributions that reflect substantial recent computation while attenuating delayed or stale information.
These mechanisms highlight that asynchrony can be realized not only through flexible aggregation schedules but also by prioritizing updates based on their timeliness and relevance.

Complementing algorithmic approaches, system-level architectures can further enable asynchronous aggregation at scale. The Kafka-ML FL framework~\cite{Towards2024Towards}, for example, leverages a distributed message-queue infrastructure in which clients push updates to the server whenever new data become available, while the server continuously aggregates incoming contributions. This decoupling of communication from computation eliminates blocking, improves throughput, and supports large-scale deployments in streaming and IoT environments.

The methods in this category are particularly effective in heterogeneous and streaming environments with intermittent client participation, as it relies on controlling update staleness and client imbalance through adaptive weighting or scheduling~\cite{reviewer1}. However, faster or data-rich clients may disproportionately influence the global model, potentially introducing bias, while stale updates can degrade convergence if not properly mitigated.

\paragraph{(2) Privacy-preserving continual release}

Privacy-preserving continual release reformulates federated aggregation as a cumulative update sharing process under formal privacy guarantees, where models release privatized summaries of historical updates instead of raw gradients at each step.

In~\cite{Cai2024Federated} the authors introduce the FL-DPCR framework, which models federated optimization as a Differentially Private Continuous Release (DPCR) process. Instead of perturbing gradients at every iteration, FL-DPCR releases noisy cumulative updates, thereby 
preventing the linear noise amplification inherent in DP-SGD (Differentially Private Stochastic Gradient Descent). A key theoretical result, the \emph{Equivalent Aggregation 
Theorem}, shows that aggregating cumulative DPCR-based local models yields the same global update as standard FedAvg, enabling 
continual model release while maintaining stable privacy--utility behavior.

To further improve release accuracy under fixed privacy budgets, the AuBCR model proposed by Cai et al.~\cite{Cai2024Boosting} introduces a dual-release mechanism based on structured matrix transformations. By organizing cumulative updates using a Binary Indexed Tree representation and an augmented strategy matrix, the method enables fine-grained allocation of privacy noise while reducing the effective dimensionality of the optimization problem. This design supports continual model release with significantly lower cumulative noise compared to standard differentially private methods, thereby improving long-term utility under strict privacy constraints.

Expanding this paradigm,~\cite{HU2025} introduces FGS-FL, which combines a \emph{Gradient Stream Release} (GSR) strategy with flat-gradient optimization. 
The GSR strategy incrementally releases aggregated gradient streams, effectively mitigating noise accumulation during training and improving stability.
When combined with Sharpness-Aware Minimization (SAM), the framework guides the model toward flatter regions of the loss landscape, reducing sharp gradients and enhancing generalization.

The methods in this category are particularly suitable for applications requiring continuous model release under strict privacy guarantees, such as healthcare or streaming analytics, because cumulative mechanisms reduce noise growth and enable long-horizon sharing. However, these methods rely on careful control of noise accumulation and often introduce additional computational overhead, while the inherent privacy--utility trade-off may become more pronounced under non-IID and drifting data distributions.

\paragraph{(3) Compression and representation-efficient aggregation}

Compression and representation-efficient aggregation reduce communication overhead by transmitting compact or partial model updates, either through parameter compression or by exchanging low-dimensional representations instead of full models.
For instance, CFedSI integrates bidirectional compression with error compensation to stabilize long-horizon training under aggressive communication reduction~\cite{Zhang2023FedSI}. 

An alternative is representation-level exchange: FedStream (TSMC)~\cite{Mawuli2023} aggregates compact prototypes instead of full parameters, and HFedMS~\cite{Zeng2023Hfedms} updates classifier layers more frequently than large feature extractors via layer-wise synchronization, often combined with semantic compression.

These methods are particularly effective in bandwidth-constrained environments, as they rely on compact updates that reduce communication without requiring full model synchronization. However, this efficiency depends on the assumption that compressed or abstracted representations retain sufficient information for learning; otherwise, information loss and delayed synchronization may accumulate over time, especially under severe heterogeneity and rapid data drift.

\paragraph{(4) Resource- and budget-aware orchestration}

Resource- and budget-aware orchestration treats aggregation as a resource-constrained optimization process, where communication, computation, and storage budgets jointly determine when and how model updates are performed. In streaming and heterogeneous environments, system resources fluctuate over time, making static aggregation schedules inefficient. This necessitates adaptive response mechanisms that can dynamically adjust to varying degrees of environmental change and resource variability~\cite{reviewer2}.

Jin et al.~\cite{Jin2021} address this challenge through the Budget-Aware Online Control (BAOC) framework, which formulates FL as a long-horizon control problem under a global cost budget. By dynamically adjusting both the number of local updates and the aggregation schedule according to system conditions such as bandwidth and latency, BAOC demonstrates that aggregation frequency itself can be treated as an adaptive control variable rather than a fixed hyperparameter.

Liu et al.~\cite{LiuWeijie2024} extend this perspective with the DYNAMITE framework, which jointly optimizes mini-batch size and aggregation frequency to balance convergence accuracy, training time, and overall cost. By adapting these parameters to device capabilities and workload dynamics, the method enables efficient operation across both static and streaming scenarios.

Recent works further generalize this perspective by explicitly incorporating data acquisition and client selection into the resource-aware optimization process. For instance, online scheduling frameworks such as TOEFL~\cite{Zhu2023} optimize data acquisition decisions in cloud-edge environments to balance learning performance and system cost, while Lyapunov-based client selection strategies such as LCCS~\cite{Su2024} dynamically regulate client participation under communication constraints. Taken together, these approaches suggest that resource-aware FCL extends beyond aggregation scheduling, requiring joint control over \emph{when}, \emph{what}, and \emph{who} contributes to the learning process.

Resource constraints also arise from limited on-device storage. Wang et al.~\cite{WangHeqiang2024} investigate this issue in streaming FL settings where clients maintain only small local caches. They propose adaptive cache management strategies that selectively retain samples to better approximate long-term data distributions. By maintaining more representative local datasets despite memory limitations, these policies improve the quality of local updates used for aggregation.

This class of methods is particularly effective in dynamic environments with fluctuating system resources, where adaptive scheduling can exploit variability in computation, bandwidth, and storage. However, their effectiveness relies on accurately modeling resource conditions; otherwise, inaccurate estimation may lead to suboptimal decisions, increased system complexity, or inefficient resource utilization.

Aggregation and Communication schemes fundamentally shape the system-level robustness of FCL: asynchrony improves availability, privacy-preserving release controls long-term information leakage, compression reduces bandwidth requirements, and orchestration adapts to dynamic resource constraints. In practice, combining these mechanisms is often necessary to sustain stable and efficient CL under real-world heterogeneity and data drift.

\subsection{Personalization Method}
Operationally, a component is considered \textit{personalized} if it is not globally aggregated and remains local to each client across tasks and time. 
Accordingly, under the Personalization axis, a method is categorized based on the client-specific component that remains persistently local and plays the primary role in handling statistical or task-level heterogeneity.
Typical examples include private classifier heads, adapters, client-specific learning rates, and local memory modules.
Explicit personalization in FCL can be viewed as a strategy for addressing persistent client heterogeneity by allowing certain parameters or modules to evolve independently from the shared model.

Overall, this paradigm shifts the learning objective from enforcing global consistency to balancing shared knowledge and client-specific adaptation.

\paragraph{(1) Optimization- and parameter-level personalization}
The cTD-$\alpha$MAML framework proposed by Li et al.~\cite{Li2023per} provides one of the clearest formulations of explicit personalization for task-incremental biometrics. It merges meta-learning with a learnable client-dependent learning rate, allowing each participant to adjust its update dynamics based on biometric modality and task evolution. The meta-optimized initialization enables fast adaptation to new incremental tasks, while the personalized learning-rate network preserves local task identity over time. This explicit control over the adaptation trajectory makes cTD-$\alpha$MAML particularly effective under heterogeneous biometric streams.

A complementary strategy is proposed by Zhang et al.~\cite{electronics14152945} through DPAO-PFL, a personalization method rooted in parameter decomposition. The global model is partitioned into shared and client-specific subsets, with the latter governed by a dynamic Fisher-information-based importance estimator. Parameters deemed essential for individual clients are shielded from global overwriting, enabling continual personalization as tasks evolve. This decomposition yields a principled mechanism for balancing global consensus and local individuality, especially when client domains diverge significantly.

From a broader perspective, optimization- and parameter-level personalization methods are particularly suitable for scenarios with strong and persistent client heterogeneity, where uniform global updates fail to capture diverse learning dynamics.
They rely on the core assumption that client-specific behaviors can be effectively modeled through personalized optimization variables or parameter partitions that remain stable across tasks.
However, this design introduces additional storage and coordination overhead, while excessive parameter isolation may limit knowledge sharing.

In summary, these methods offer fine-grained control over client adaptation, but require careful balancing between personalization and collaboration to remain effective in large-scale federated settings.

\paragraph{(2) Feature- and head-level personalization}

Feature- and head-level personalization introduces a structural separation between shared representations and client-specific decision layers, enabling flexible adaptation without modifying the entire model.

In domain-heterogeneous speaker recognition, Chen et al.~\cite{chen2023learning} extend explicit personalization to the feature space by introducing a client-dependent projection layer on top of a federated Transformer backbone. This private projection module maps shared embeddings into a domain-adapted latent representation, capturing client-specific acoustic variability (e.g., room conditions and microphone differences). Since this module is never shared with the server, it provides a structurally isolated pathway for continual personalization as local speech data evolve over time.

Relatedly, head-level personalization keeps the representation extractor globally shared while allowing each client to maintain a private classifier that evolves across tasks. The Meta-RBCIL framework proposed by Zheng et al.~\cite{Zheng2025Meta} separates a globally shared representation extractor from a personalized classifier maintained locally by each client. The shared representation is meta-trained to remain stable over incremental tasks, while the client-specific classifier adapts to class-incremental dynamics without overwriting global knowledge. This dual-branch design provides explicit and persistent personalization, enabling clients to specialize their decision boundaries while still benefiting from federated feature learning.

These methods are well-suited for domain-heterogeneous settings where shared feature representations can be combined with personalized decision layers.
They assume that the global representation remains sufficiently expressive and transferable across clients.
However, under strong distribution shifts, feature–head mismatches may degrade performance, and maintaining private components increases local complexity.
Overall, they provide a modular approach to balancing shared learning and local adaptation, but depend on the quality of the learned feature space.

\paragraph{(3) Adapter- and module-based personalization}

Adapter- and module-based personalization reformulates CL as a modular decomposition problem, in which shared backbones capture general knowledge while lightweight client-specific modules encode task- or domain-specific variations.
This modular design is 
particularly attractive in FCL because it allows clients to retain 
task-sensitive knowledge locally while still benefiting from global shared representations. 

SacFL~\cite{Zhong2025} adopts a principled separation between a \textit{task-robust encoder} shared across all clients and 
\textit{task-sensitive lightweight decoders} that remain entirely local. This structural split enables continual personalization 
without inflating communication costs: only the shared encoder is federated, whereas each client maintains its own decoder to 
capture domain-, task-, or distribution-specific nuances.
To ensure adaptability under streaming or adversarial drift, SacFL integrates a contrastive-learning-based drift detector and 
a degradation-aware adversarial monitoring module. 

A key advantage of such designs is their ability to provide implicit regularization across tasks: the shared encoder gradually stabilizes through joint optimization, while lightweight local modules specialize with minimal interference and computation.
This property is particularly beneficial for 
resource-constrained devices, where memory and compute budgets impose strict limits on model updates.

Although the Adapter--Retainer framework was previously discussed as a structure-aware adaptation strategy, it can also be interpreted as an explicit personalization mechanism. In particular, the client-specific adapters evolve locally, whereas the retainer preserves globally shared knowledge.

The Adapter--Retainer framework proposed by Yang et al.~\cite{YANG202416} introduces a modular decomposition of the global model into two complementary components: a client-specific \emph{Adapter} and a shared \emph{Retainer}. The Adapter, denoted by $A_i$ for client $i$, captures task-specific knowledge using the local dataset $\mathcal{D}_i$, while the Retainer $R$ preserves long-term historical representations and remains fixed on the server. Formally, the global model is represented as
\begin{equation}
    M = [A, R],
\end{equation}
where $A$ and $R$ denote the Adapter and Retainer parameters, respectively, and $[\cdot,\cdot]$ indicates their composition. During training, only the Adapter is updated locally on each client, and the resulting client-specific Adapters are aggregated at the server to form an updated Adapter, while the Retainer remains unchanged. The updated model is then reconstructed by combining the aggregated Adapter with the shared Retainer.
This separation allows clients to specialize to their local task distributions through the Adapter, while leveraging the Retainer as a stable knowledge backbone, thereby mitigating catastrophic forgetting with reduced computational overhead.

In a related but more task-granular design, FCCL~\cite{PanWang2025FCCL} introduces \emph{lightweight adapters} attached to a 
shared backbone. For each new task, clients instantiate a new adapter, a small parameter block, and use a data-free task 
discriminator to automatically select the appropriate adapter during inference. The shared backbone learns generic 
representations, while adapters capture incremental, task-specific variations.
The resulting architecture enables high scalability in CL settings: as the number of tasks grows, only small 
adapters accumulate, significantly reducing storage overhead compared with full model copies or replay-based strategies. 
Moreover, the task discriminator allows FCL systems to dynamically select the most relevant adapter even under heterogeneous 
and non-IID client environments.

From a broader perspective, this paradigm is particularly suitable for continual and resource-constrained environments, where isolating task-specific knowledge in lightweight modules reduces interference and communication overhead.
It relies on the core assumption that a shared backbone can learn stable and transferable representations across tasks, while modular components can effectively capture task-specific variations.
However, as the number of tasks grows, the accumulation of adapters may increase memory and system complexity, and selecting the appropriate module during inference introduces additional computational overhead.

In summary, adapter- and module-based personalization provides a scalable and modular solution for balancing stability and plasticity, but requires careful management of module growth and selection mechanisms.

\paragraph{(4) Task- and memory-grained personalization}
 
Beyond adapter-based designs, a more structured form of personalization has recently emerged in FCL, where adaptation is organized at the level of tasks, memories, or hierarchical roles. Rather than maintaining a single personalized subnetwork, these approaches allocate task-specific memories or structured modules that enable fine-grained knowledge sharing across heterogeneous clients.

Cross-FCL~\cite{CrossFCL2024} introduces task-level personalization by allowing clients to selectively distill knowledge from previously learned tasks whose representations are semantically similar to the current objective. Instead of performing global model averaging, the framework activates a similarity-based task-sharing mechanism, reducing negative transfer under heterogeneous task trajectories.

Hierarchical personalization is explored in CFLViT~\cite{CFLViT2025}, where clients maintain multiple model roles (e.g., Junior, Consultant, Senior Consultant) responsible for short-, medium-, and long-term adaptation. This layered design enables domain-specific specialization while preserving shared high-level representations within a federated backbone.

In industrial environments, MeCo~\cite{Chen2025know} adopts a meta-knowledge perspective by constructing a server-side knowledge pool indexed by task embeddings. Clients dynamically retrieve relevant knowledge components based on similarity to their current task, enabling flexible composition of personalized models without exchanging full parameters.

An even more explicit task-grained formulation is proposed in Loci~\cite{Luopan2025Loci}, which replaces global aggregation with retrieval from a hierarchical task-memory structure. Clients contribute compact summaries of completed tasks, and subsequent training retrieves only semantically aligned components. This selective aggregation mechanism substantially reduces communication while improving robustness in heterogeneous task settings.

From a broader perspective, these methods are particularly suitable for highly heterogeneous and evolving task environments, where fine-grained knowledge sharing based on task similarity is essential.
They rely on the core assumption that task relationships can be accurately modeled and that relevant knowledge can be effectively retrieved from structured memories or task pools.
However, this design increases system complexity and depends heavily on reliable similarity estimation; as the number of tasks and clients grows, it may also reduce global knowledge sharing and lead to fragmentation.

In summary, task- and memory-grained personalization enables flexible and selective knowledge transfer, but requires careful design of similarity metrics and memory structures to remain scalable and effective.

\subsection{CL Paradigm}

Unlike the previous axes, which categorize methods by their algorithmic mechanisms, this axis is scenario-driven and classifies FCL approaches according to the type of CL problem they address.
Operationally, under the CL Paradigm axis, a method is categorized according to the form of non-stationarity assumed in the problem setting, rather than according to the optimization or architectural mechanism used to solve it.
Therefore, the methods mentioned here serve only as illustrative examples of each paradigm, while their algorithmic details are discussed in the Adaptation, Aggregation, and Personalization sections.

\paragraph{(1) Task-incremental FCL}
Task-incremental FCL models federated systems that encounter a sequence of \emph{explicitly defined tasks}, where each task is associated with its own dataset and a known task identifier. The objective is to learn each incoming task while retaining performance on previously learned ones. This setting is common in dynamic environments such as Mobile Edge Computing (MEC) and industrial monitoring, where evolving operational conditions naturally create successive task segments and shift local data distributions~\cite{Li2023per}.

A defining assumption is that task identity is available during both training and inference, enabling explicit task-conditioned behavior (e.g., selecting the appropriate head, adapter, or routing path). Consequently, the main difficulty is not recognizing tasks, but managing the stability--plasticity trade-off under federated constraints: clients may follow different task orders, exhibit non-IID data even within the same task, and participate intermittently with heterogeneous compute and communication budgets~\cite{yang2024federated}.

Representative solutions span multiple mechanism families: sparse adaptation for resource-constrained edge devices~\cite{Liu2025Spars}, architectural modularity for reducing cross-task interference~\cite{zhou2025cross}, meta-learning for fast task adaptation under heterogeneous clients~\cite{Li2023per}, replay-enhanced stabilization across task transitions~\cite{LI2024111491,SHAMI2025114976}, and adapter-based designs that isolate task-specific changes with minimal additional parameters~\cite{PanWang2025FCCL}. Despite their diversity, these methods share the same premise: \emph{task boundaries are explicit}, so the core objective is continual retention and controlled transfer across a sequence of known tasks under federated constraints.

In evaluation, task-incremental FCL is commonly assessed using (i) average accuracy over all previously observed tasks and (ii) forgetting measures that quantify performance degradation on earlier tasks after learning subsequent ones. In federated environments, these metrics are further influenced by partial participation and cross-client heterogeneity, since clients may contribute updates at different frequencies and may not share identical task sequences.

This paradigm is particularly suitable for scenarios with clearly defined task boundaries and explicit task identities.
It relies on the core assumption that task-specific components can be reliably selected during both training and inference.
However, in real-world federated deployments where task boundaries are ambiguous or client task sequences are misaligned, this assumption may break down, leading to coordination challenges and potential negative transfer.

Overall, task-incremental FCL provides a structured and interpretable setting for CL, but its effectiveness depends on the clarity and consistency of task definitions across clients.

\paragraph{(2) Class-incremental FCL}

Class-incremental FCL considers scenarios where new classes are introduced over time without explicit task boundaries, requiring models to continuously expand their label space while retaining knowledge of previously learned classes.

In contrast to task-incremental settings, where task identities are explicitly available, class-incremental FCL assumes a \emph{shared and progressively expanding output space}. At each stage, clients are exposed only to a subset of classes, and previously observed class data may no longer be accessible due to privacy, storage, or communication constraints. As a result, models must learn to recognize newly introduced classes while maintaining performance on earlier ones, without relying on task identifiers during inference.

This setting naturally arises in federated environments with evolving label spaces, such as emerging diseases in healthcare, new attack types in cybersecurity, or expanding object categories in distributed vision systems. In such scenarios, clients may introduce classes asynchronously, leading to severe label skew and non-IID distributions.

A defining challenge in class-incremental FCL is the combined effect of \emph{catastrophic forgetting} and \emph{class imbalance} under federated constraints. Since historical data from previous classes are often unavailable, models tend to bias toward recently introduced classes. This issue is further amplified by partial client participation and heterogeneous data distributions, making it difficult to maintain consistent decision boundaries across all observed classes.

Representative solutions span several mechanism families. Adapter-based approaches, such as FCCL~\cite{PanWang2025FCCL}, isolate newly learned knowledge via lightweight modules while preserving shared representations, enabling scalable expansion with reduced forgetting. Replay-based methods leverage generative models to approximate past data distributions; for example, diffusion-driven frameworks such as FedCapD~\cite{Iqbal2026} and weakly supervised diffusion-based replay methods~\cite{Sun2025Federated} generate synthetic samples of earlier classes to stabilize training under non-IID conditions.

Application-driven studies further highlight the relevance of this paradigm. In cybersecurity, federated class-incremental learning has been applied to encrypted traffic classification~\cite{ID1} and intrusion detection systems~\cite{jin2024fliids}, where new categories continuously emerge and must be incorporated without retraining from scratch. More advanced approaches integrate personalization and adaptive mechanisms: SacFL~\cite{Zhong2025} combines class-incremental learning with drift-aware adaptation and resource-efficient model design, while meta-learning-based methods such as Meta-RBCIL~\cite{Zheng2025Meta} improve generalization and reduce forgetting through balanced representation learning across clients.

In evaluation, class-incremental FCL is typically assessed using (i) overall accuracy across all observed classes and (ii) forgetting measures that quantify performance degradation on earlier classes after new classes are introduced. In federated settings, additional factors such as class imbalance, client heterogeneity, and communication efficiency further influence these metrics.

From a broader perspective, this paradigm is particularly suitable for scenarios with evolving label spaces and no explicit task boundaries.
It relies on the core assumption that models can incrementally expand their classification capacity while maintaining a unified prediction space across all classes.
However, the absence of task identifiers and limited access to historical data significantly increase the difficulty of preventing forgetting and mitigating bias toward recent classes.

Overall, class-incremental FCL provides a realistic and widely applicable formulation, but requires effective mechanisms for memory retention, class balancing, and scalable model adaptation under federated constraints.

\paragraph{(3) Domain-incremental and cross-domain FCL}

In contrast to task-incremental settings, domain-incremental and cross-domain FCL consider scenarios in which the \emph{task remains conceptually the same} while the underlying data distribution changes across domains, institutions, acquisition conditions, or modalities. Rather than switching between distinct tasks, clients experience domain drift, requiring models to continuously adapt to new environments while maintaining performance on previously encountered domains~\cite{chen2023learning}.

This setting is especially common in federated deployments where data heterogeneity arises from real-world variability across devices or locations. Because domain identity may be implicit or only partially observable, the central challenge is to achieve robustness to distribution shifts without catastrophic forgetting, often under non-IID and asynchronously evolving conditions.

Representative works illustrate the breadth of this paradigm. Cross-FCL~\cite{CrossFCL2024} explores cross-edge adaptation, enabling models to transfer knowledge across domains associated with heterogeneous operational contexts. In multi-domain medical imaging, CFLViT~\cite{CFLViT2025} adopts a hierarchical continual learning strategy across datasets such as OCT, MedPix, and NIH Chest X-ray, allowing a shared ViT backbone to accommodate domain-specific visual variations while preserving common semantic features.

Domain-incremental challenges are equally prominent in audio and physiological applications. The personalized FL speaker-recognition framework proposed by Chen et al.~\cite{chen2023learning} models gradual shifts caused by acoustic environments, language differences, and recording devices, demonstrating how domain adaptation naturally arises in user-centric systems. MetaCL~\cite{SUN2023586} addresses time- and institution-dependent drift in physiological signals, where underlying biomedical processes evolve across populations and measurement conditions. Extending this perspective, F-FCL~\cite{IQBAL2025107920} introduces a family-based continual learning mechanism for multi-domain analysis spanning vision, sensor, and textual data, highlighting the need for flexible knowledge transfer across heterogeneous modalities.

From a broader perspective, this paradigm is particularly suitable for real-world federated applications with implicit and continuous distribution shifts.
It relies on the core assumption that models can learn domain-invariant or adaptable representations across clients.
However, the absence of explicit task identifiers complicates adaptation and increases the risk of gradual performance degradation without effective drift-aware mechanisms.

In summary, domain-incremental FCL provides a realistic and practical formulation, but its effectiveness depends on robust representation learning under continuous distribution shifts.

\paragraph{(4) Online, streaming, and drift-oriented FCL}

A more demanding paradigm arises when data arrive continuously with no clear boundaries between tasks or domains. In such online and streaming environments, models must adapt to concept drift in real time while operating under the communication, privacy, and resource constraints of FL. Often referred to as \textit{streaming FCL}, this setting has become increasingly important due to the proliferation of edge devices that generate high-velocity data streams.

Unlike task- or domain-incremental scenarios, streaming FCL assumes that distribution changes are continuous and potentially unpredictable. Consequently, the central challenge is to achieve rapid plasticity for new information while preserving useful historical knowledge, without relying on explicit task identifiers or batch retraining.

Early work by Nandi and Xhafa~\cite{Nandi2023DockerFederated,NANDI2022340} demonstrates an interleaved test-then-train online federated framework for real-time multimodal emotion recognition, highlighting the feasibility of continual adaptation in latency-sensitive applications. Drift-aware extensions such as CDA-FedAvg~\cite{Casado2022ConceptDrift} integrate concept drift detection with rehearsal-based updates, enabling the global model to remain stable as data distributions evolve.

A broader family of streaming methods has emerged under the FedStream framework. The TNSE variant~\cite{WangNaiyu2024}, based on the $H_{\text{str}}$SAGA optimizer, addresses heterogeneity in both data distributions and communication patterns, while the TSMC variant~\cite{Mawuli2023} employs prototype-based aggregation to maintain robustness under distribution shifts. FedMuL~\cite{Lamptey2025} extends this approach to multilabel data streams, emphasizing that continual federated adaptation must handle both concept drift and evolving label spaces.

In parallel, a growing body of application-driven work demonstrates the effectiveness of streaming FCL under real-world concept drift. For example, adaptive federated frameworks for tasks such as gender classification~\cite{Sharma2025}, mobile malware detection~\cite{Augello2025}, and load forecasting in smart grids~\cite{Elgalhud2025} show that integrating drift-aware updates and continuous adaptation significantly improves robustness in non-stationary environments. Similarly, recent approaches such as DISFIDA~\cite{Gelenbe2024} combine online learning with self-supervised objectives to enable scalable and data-efficient intrusion detection in IoT systems.

Theoretical analyses further illuminate this paradigm. Streaming FL Cache Rules~\cite{WangHeqiang2024} examine how limited on-device storage and buffering strategies influence long-term learning dynamics. 
Recent advances further strengthen both the theoretical and practical foundations of streaming FCL by explicitly modeling time-varying environments and adversarial conditions. For instance, online federated optimization frameworks such as BR-OMGD~\cite{Tian2026} establish dynamic regret guarantees under non-IID and Byzantine settings, demonstrating that FCL can be grounded in principled online optimization theory.

Complementary algorithmic solutions include FedARF~\cite{Massaoudi2025}, which incorporates drift-detection ensembles such as ADWIN and DDM, and SOFed~\cite{Shi2023Self}, a self-supervised approach designed for unlabeled high-throughput streams.

Semi-supervised frameworks such as SFLEDS~\cite{Cobbinah2023} employ drift-aware prototype refinement, while modular approaches like Adapter--Retainer~\cite{YANG202416} use lightweight, continually updated adapters to capture new information without overwriting shared knowledge. 

Together, these methods demonstrate that streaming FCL requires coordinated mechanisms for rapid adaptation, memory preservation, and communication efficiency.
Moreover, communication-efficient online knowledge transfer mechanisms, such as federated distillation-based approaches~\cite{Salman2026}, further extend streaming FCL by enabling continual adaptation without explicit parameter sharing, highlighting a shift toward more scalable and privacy-aware designs.

From a broader perspective, this paradigm is particularly suitable for real-time and high-velocity data environments.
It relies on the core assumption that models can continuously adapt to evolving distributions without explicit task boundaries.
However, persistent concept drift, limited supervision, and strict resource constraints significantly increase the difficulty of maintaining stability, and failures in drift handling can lead to rapid performance degradation.
In summary, streaming FCL offers maximum flexibility in dynamic environments, but requires robust drift-aware and resource-efficient learning mechanisms.

\begin{landscape}

\begin{figure}
    \centering
    \begin{subfigure}{0.65\textwidth}
        \includegraphics[width=\linewidth]{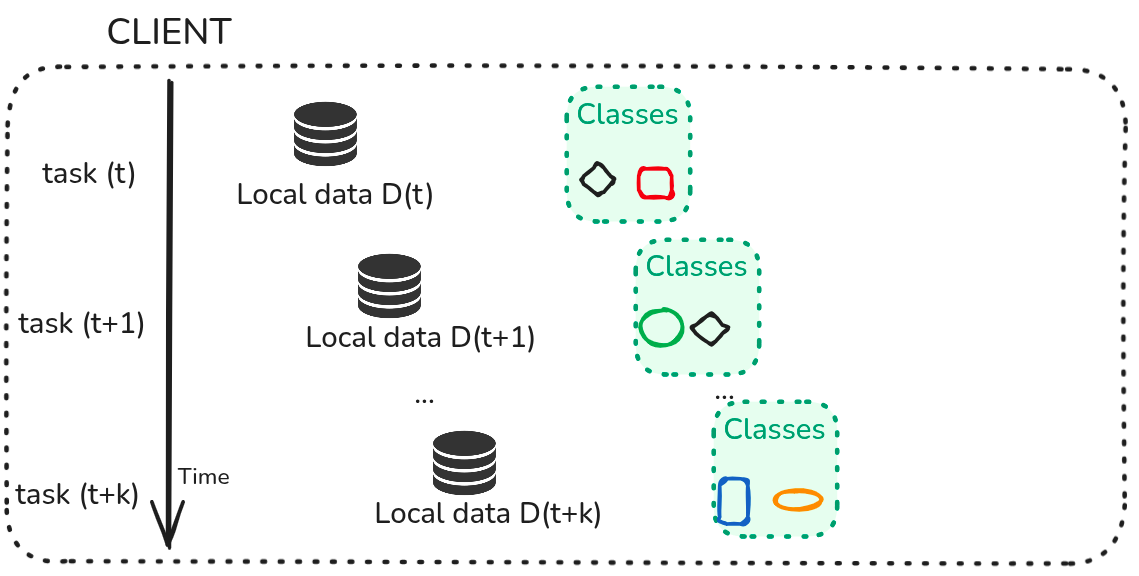}
        \caption{Task incremental}
    \end{subfigure}
    \begin{subfigure}{0.65\textwidth}
        \includegraphics[width=\linewidth]{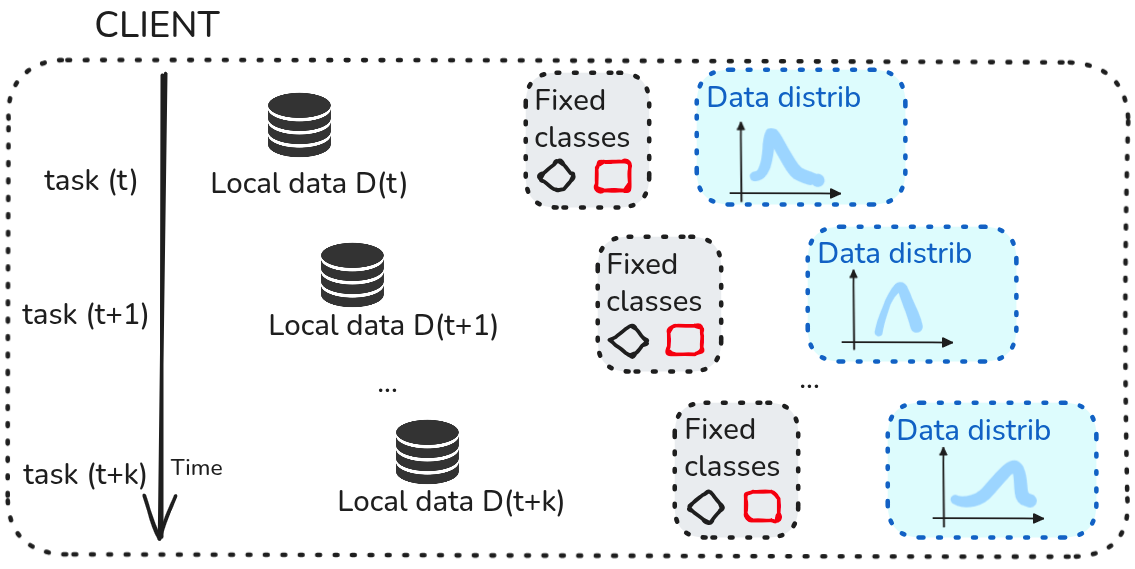}
        \caption{Domain Incremental}
    \end{subfigure}\\ \vspace{1cm}
    \begin{subfigure}{0.65\textwidth}
        \includegraphics[width=\linewidth]{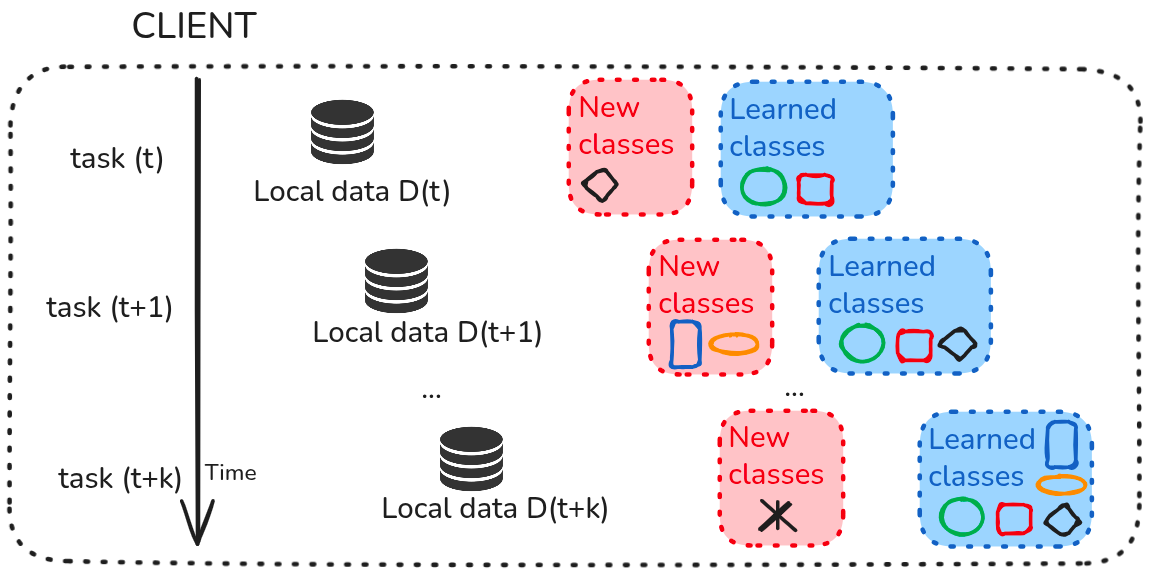}
        \caption{Class incremental}
    \end{subfigure}
    \begin{subfigure}{0.65\textwidth}
        \includegraphics[width=\linewidth]{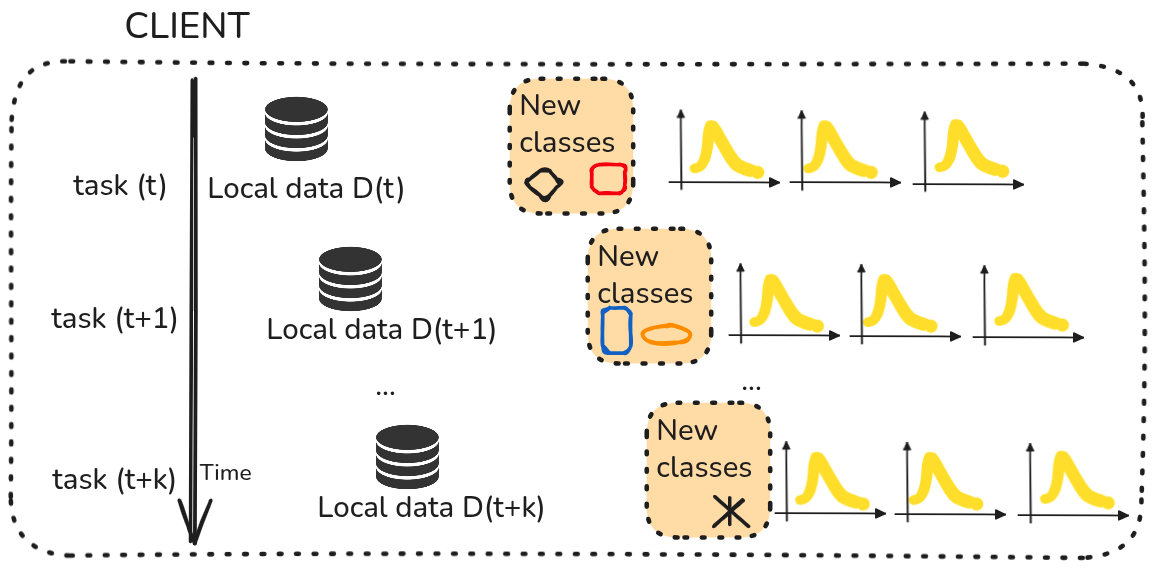}
        \caption{Streaming}
    \end{subfigure}
    \caption{Comparison of different CL scenarios in federated settings. (a) Task-incremental learning, where data are organized into sequential tasks with task-specific distributions.  (b) Domain-incremental learning, where the class space remains fixed while the data distribution evolves over time. (c) Class-incremental learning, where new classes are progressively introduced and previously learned classes must be retained. (d) Online (streaming) learning, where data arrive continuously over time, potentially involving evolving distributions and class sets.}
    \label{fig:CL_paradigms}
\end{figure}

\end{landscape}

This axis highlights that the choice of CL paradigm fundamentally shapes the problem formulation in FCL.
Task-incremental settings assume explicit boundaries, domain-incremental scenarios focus on continuous distribution shifts, and streaming paradigms address fully unstructured data evolution.
Understanding these distinctions is essential for selecting appropriate adaptation, aggregation, and personalization strategies in practical federated deployments.
A compact visualization of the main differences between these paradigms is provided in Figure \ref{fig:CL_paradigms}.

As an application of the proposed taxonomy, which organizes the previously fragmented body of FCL research into a coherent design space, Tables~\ref{tab:FCL_taxonomy_part1}  report some of the most significant methods, classified exploiting the taxonomy. 
Each method is labeled separately along the Adaptation, Aggregation and Communication, and Personalization axes according to its dominant algorithmic mechanism, while the CL Paradigm axis specifies the underlying non-stationary learning scenario addressed by the method. This axis-wise representation enables consistent alignment between the taxonomy, the tables, and the textual discussion, while preserving the multi-dimensional nature of practical FCL systems. It also enables researchers and practitioners to analyze trade-offs, identify gaps, and systematically design new FCL algorithms by combining complementary mechanisms across multiple axes.

\begin{landscape}
\begin{table*}[p]
\centering
\caption{Representative FCL methods organized according to the proposed taxonomy (Part I). }
\label{tab:FCL_taxonomy_part1}
\scriptsize
\renewcommand{\arraystretch}{1.25}
\begin{tabular}{p{2.0cm} p{2.6cm} p{2.3cm} p{3.4cm} p{4.6cm} p{3.6cm}}
\hline
\textbf{Primary axis (dominant mechanism)} & \textbf{Method} & \textbf{CL paradigm} & \textbf{Main challenge} & \textbf{Core idea / solution} & \textbf{Limitations} \\
\hline

Adaptation &
FCL-BL~\cite{Le2021} &
Task-incremental, asynchronous FL &
Catastrophic forgetting under asynchronous and streaming clients &
Incremental broad-learning updates combined with batch-asynchronous aggregation for stable continual FL &
Limited to BL architectures; difficult to scale to deep neural models \\

Adaptation &
FedSI / CFedSI~\cite{Zhang2023FedSI} &
Class-/task-incremental &
Forgetting in highly non-IID data and high communication cost &
SI-based parameter-importance regularization with bidirectional compression and error compensation &
Requires importance statistics and error buffers; sensitive to compression hyperparameters \\

Adaptation &
FINAL~\cite{Zhao2024FINAL} &
Generic continual FL &
Long-term knowledge retention under non-IID clients &
Federated Fisher matrix aggregation and Fisher-based regularization to preserve critical parameters &
High Fisher computation and communication overhead; limited scalability to large models \\

Adaptation &
FCL4DD~\cite{Sun2025Federated} &
Class-incremental medical imaging &
Absence of historical data under strict privacy constraints &
Weakly supervised diffusion-based generative replay for reconstructing past samples &
High training cost of diffusion models; replay quality depends on generator fidelity \\

Adaptation &
FedCapD~\cite{Iqbal2026} &
Class-incremental FCL (non-IID) &
Catastrophic forgetting under non-IID data without raw data access &
Diffusion-based generative replay with capsule distillation for privacy-preserving knowledge retention &
High computational complexity; challenging deployment on resource-constrained devices \\

Adaptation &
TRINA~\cite{LI2024111491} &
Task-/regression-incremental IIoT &
Spatiotemporal drift in industrial edge streams &
Self-challenge rehearsal combined with gradient balancing for robust continual adaptation &
Complex pipeline; domain-specific tuning required \\

Adaptation &
Sparse-FCL~\cite{Liu2025Spars} &
Task-incremental MEC &
Resource constraints and task interference on edge devices &
Dynamic sparse training with stable–plastic neuron separation and task-adaptive topology evolution &
Performance depends on sparsity scheduling; hardware-aware sparsity needed \\

Aggregation &
FedStream (TNSE)~\cite{WangNaiyu2024} &
Streaming / online CL &
Dual heterogeneity in data distributions and traffic patterns &
Asynchronous weighted aggregation based on computation effort, data size, and staleness &
Potential client bias; sensitive to staleness-weighting design \\

Aggregation &
FedStream (TSMC)~\cite{Mawuli2023} &
Streaming / online CL &
Concept drift under bandwidth constraints &
Prototype and microcluster-based aggregation with compressed representations &
Prototype compression may lose fine-grained information \\

Aggregation &
Kafka-ML FL~\cite{Towards2024Towards} &
Streaming FL in IoT &
Irregular client arrivals and large-scale asynchronous orchestration &
Kafka-based asynchronous FL with topic-driven scheduling and continuous aggregation &
Dependent on Kafka infrastructure; increased system complexity \\

Aggregation &
FL-DPCR \& AuBCR~\cite{Cai2024Federated,Cai2024Boosting} &
Continual FL with differential privacy &
Privacy--utility degradation in long-horizon training &
Differentially private continuous release using optimized matrix mechanisms with BIT &
Accuracy–privacy trade-off remains; computation overhead from DP matrix operations \\

\hline
\end{tabular}
\end{table*}
\end{landscape}

\begin{landscape}
\begin{table*}[p]
\centering
\caption{Representative FCL methods organized according to the proposed taxonomy (Part II, continued).}

\label{tab:FCL_taxonomy_part2}
\scriptsize
\renewcommand{\arraystretch}{1.25}
\begin{tabular}{p{2.0cm} p{2.6cm} p{2.3cm} p{3.4cm} p{4.6cm} p{3.6cm}}
\hline
\textbf{Primary axis (dominant mechanism)} & \textbf{Method} & \textbf{CL paradigm} & \textbf{Main challenge} & \textbf{Core idea / solution} & \textbf{Limitations} \\
\hline

Aggregation &
FGS-FL~\cite{HU2025} &
DP streaming FCL &
Instability of noisy gradients in long-horizon DP training &
Flat-gradient optimization combined with DP gradient-stream release &
High computational overhead; limited accuracy under tight privacy budgets \\

Aggregation &
CFedSI with BCEC~\cite{Zhang2023FedSI} &
Class-/task-incremental &
High bi-directional communication cost in continual FL &
Gradient sparsification with bidirectional error compensation to stabilize compressed updates &
Requires extra memory; sensitive to sparsification ratios \\

Aggregation &
HFedMS~\cite{Zeng2023Hfedms} &
Task-/class-incremental FCL &
Communication and memory limits in heterogeneous data streams &
Layer-wise alternative synchronization with semantic compression and replay &
Semantic memory design is complex; async layer management adds overhead \\

Personalization &
cTD-$\alpha$MAML~\cite{Li2023per} &
Task-incremental biometrics &
Fast personalization under heterogeneous biometric modalities &
Meta-learned initialization and adaptive learning-rate networks with exemplar-guided distillation &
Meta-training is costly; relies on exemplar storage and careful regularization \\

Personalization &
DPAO-PFL~\cite{electronics14152945} &
Task-/class-incremental PFL &
Balancing shared and personalized parameters &
Dual parameter decomposition with Fisher-based protection of client-critical weights &
Requires accurate importance estimation; extra per-client storage \\

Personalization &
Speaker-PFL~\cite{chen2023learning} &
Domain-incremental audio FL &
Domain and acoustic variability across users &
Shared transformer encoder with private projection heads for domain adaptation &
Additional local parameters; depends on sufficient client-side data \\

Personalization &
Meta-RBCIL~\cite{Zheng2025Meta} &
Class-incremental FCL &
Maintaining a stable global representation while personalizing users &
Meta-trained backbone with lightweight client-specific classifiers &
Assumes stable feature space; limited scalability to very large label sets \\

Personalization &
SacFL~\cite{Zhong2025} &
Task-/streaming FCL under drift &
Data drift and adversarial degradation in streaming environments &
Shared robust encoder with private lightweight decoders and contrastive drift detection &
Multi-module complexity; threshold tuning required \\

Personalization &
Loci~\cite{Luopan2025Loci} &
Task-incremental heterogeneous NLP/CV/GNN &
Severe cross-task divergence and high communication overhead &
Task-memory palace with task-grained aggregation via compact task summaries &
Memory structure grows with number of tasks; similarity metric quality is critical \\

\hline
\end{tabular}
\end{table*}
\end{landscape}

\section{Application Domains and Use Cases}
\label{sec:applic}
FCL is particularly valuable in real-world settings where data are both decentralized and continuously evolving. This section examines four representative application domains in which these challenges are especially pronounced: healthcare, IIoT, smart cities, and environment. Across these domains, FCL enables models to adapt over time, retain previously acquired knowledge, and operate under strict privacy and communication constraints without centralizing raw data.

\begin{figure}[t]
    \centering
    \includegraphics[width=0.9\linewidth]{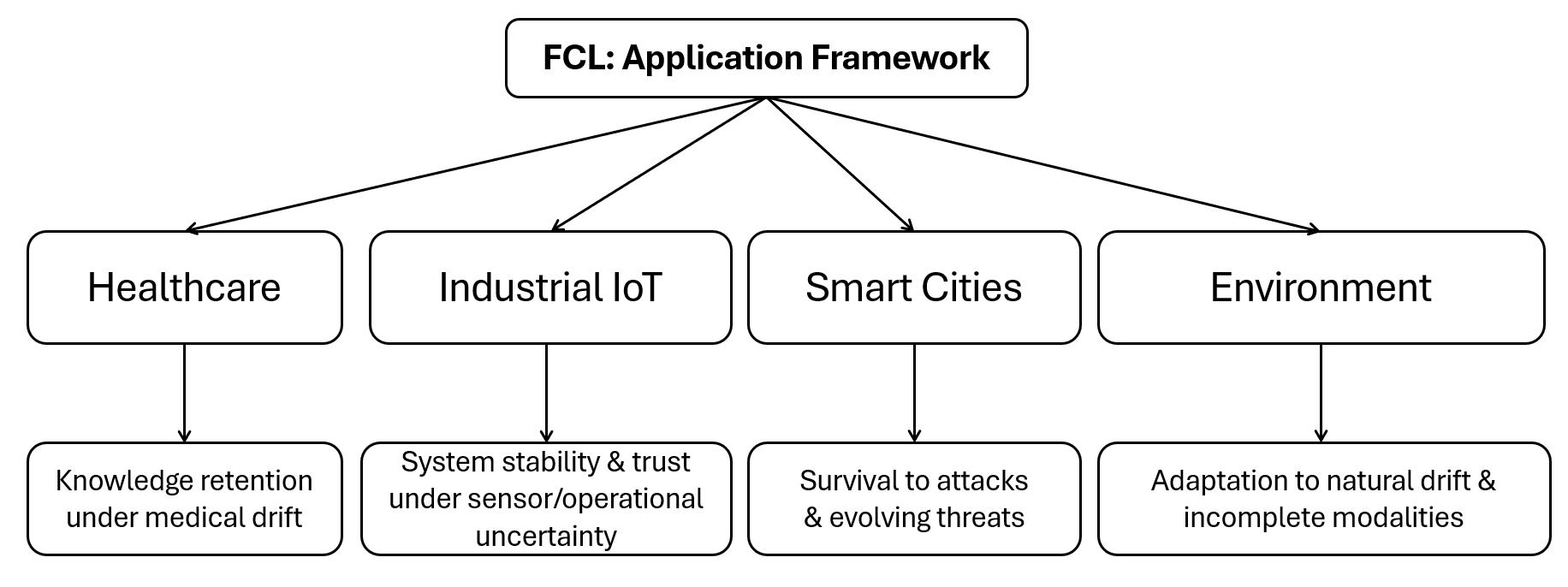}
    \caption{A conceptual view of FCL across application domains: healthcare, IIoT, smart cities, and environment.}
    \label{fig:FCL_taxonomy}
\end{figure}

This section focuses on domain-specific implementations and practical trade-offs of FCL systems, while data characteristics and evaluation protocols are discussed separately in subsequent sections.
Building on the definition of FCL introduced earlier—encompassing both explicit task-incremental learning and implicit adaptation to streaming, non-stationary data—we analyze how federated continual principles are instantiated in practical deployments. Fig.~\ref{fig:FCL_taxonomy} provides a conceptual overview of the domain-specific requirements that drive different forms of continual adaptation.

Each domain imposes distinct requirements. 
Healthcare emphasizes reliable integration of new clinical evidence while preserving prior knowledge; IIoT prioritizes robustness and efficiency under dynamic operating conditions; security requires resilience against evolving and adversarial contexts; and environment focuses on long-term adaptation in complex and heterogeneous data settings.

The following subsections review representative advances in each domain, highlighting both shared design patterns and domain-specific challenges in FCL systems.

\subsection{Healthcare}

Recent developments in FCL for healthcare reveal a convergent movement toward models capable of sustaining diagnostic performance under complex and privacy-restricted medical environments. Across diverse modalities, including thoracic imaging, dermatology, physiological signals, and biometrics, these works demonstrate that meaningful continual adaptation in federated settings requires explicit mechanisms for knowledge preservation, stable long-term adaptation, and architectural designs that can withstand the compounded effects of heterogeneity and sequential task arrival~\cite{zhou2025cross}.

In thoracic infection diagnosis, cross-paradigm integration of FL with CL has been explored through MLP-Mixer backbones enriched with Learning without Forgetting, Elastic Weight Consolidation, and auxiliary regularizers, as shown in~\cite{zhou2025cross}. These studies highlight that distillation-based knowledge preservation, particularly in the LwF variant, maintains representational stability more effectively than static parameter-importance constraints when tasks emerge asynchronously across institutions. Although absolute accuracy remains modest due to data fragmentation and class imbalance, this line of work establishes the operational viability of FL--CL hybrids in multi-center diagnostic pipelines under evolving clinical conditions. From a practical perspective, this suggests that regularization- and distillation-based FCL is more suitable for short- to medium-term model evolution, but may struggle to scale to truly lifelong clinical learning where task diversity and model capacity requirements grow continuously.

Generative replay introduces a more expressive route to continuality. The FCL4DD framework~\cite{Sun2025Federated} employs a weakly supervised diffusion model to synthesize privacy-preserving pseudo-samples that emulate earlier disease distributions, providing a robust defense against catastrophic forgetting in dermatology and other image-based diagnostic tasks. Through attention-guided conditioning and feature clustering, diffusion-based replay maintains diversity in reconstructed samples despite limited labeling, enabling the global model to integrate new disease classes without erasing prior ones. However, this benefit comes at the cost of high computational overhead and limited interpretability aligned with clinical decision-making, underscoring the tension between generative fidelity and deployability in constrained healthcare environments. This trade-off suggests that diffusion-based replay is currently better suited to offline or periodically updated clinical models than to deployment in resource-constrained hospital settings.

Complementing generative approaches, hierarchical architectures such as CFLViT, introduced by~\citep{CFLViT2025}, provide structured multi-level knowledge retention to address domain inconsistency across medical centers. By partitioning learning into short-term adaptation (Junior models), intermediate consolidation (Consultant models), and long-term retention (Senior Consultant models), these frameworks embed continuality directly into the model hierarchy rather than attaching it through regularization. Vision Transformer backbones, augmented with multi-resolution fusion, autoencoding, and distillation across hierarchy levels, further enhance domain invariance while enabling limited forms of anatomical interpretability through attention maps. Although computationally demanding, hierarchical FCL demonstrates consistent gains in accuracy and reduced forgetting across retinal, radiological, and multi-domain imaging datasets. More importantly, this design offers a structurally scalable alternative to parameter-regularization and replay-based methods by decoupling short-term plasticity from long-term memory consolidation.

At the intersection of identity recognition and healthcare-adjacent biometrics, personalized FCL extends these ideas by embedding meta-learning within federated optimization. The cTD-$\alpha$MAML approach proposed by~\cite{Li2023per} illustrates how clients can maintain personalized task trajectories while benefiting from a shared meta-initialization that accelerates adaptation to new classes. Prototype replay, task-level distillation, and per-parameter learnable learning rates collectively minimize interference between old and new biometric identities. The framework exhibits strong generalization under incremental biometric scenarios, although it remains sensitive to exemplar selection and imposes notable computational demands on edge devices. This reveals a practical tension between personalization and scalability, as highly adaptive client-specific models may become difficult to sustain in large federated healthcare networks.

Moreover, personalized continual FL for speaker recognition, as explored by~\cite{chen2023learning}, demonstrates how identity-conditioned feature modulation and personalized routing mechanisms can stabilize long-term acoustic representation learning across evolving speaker domains. However, such identity-conditioned architectures may over-specialize to client-specific patterns, reducing robustness under substantial client-level variation.

In physiological signal analysis, \cite{SUN2023586} integrates FL with blockchain-backed meta-continual optimization to ensure tamper-resistant model provenance and robust task adaptation. Their framework leverages meta-learned priors, replay buffers, and distributed-ledger-based verification to handle sequential task arrival in highly sensitive biosignal monitoring applications. While this enhances trust and traceability, the reliance on blockchain infrastructure also increases system complexity, raising practical concerns about latency and scalability.

Rather than converging on a single dominant strategy, healthcare-oriented FCL research has evolved through multiple complementary stages, ranging from parameter-regularized and distillation-driven continuality, to generative replay, hierarchical consolidation, and finally meta-personalized adaptation. Each paradigm introduces a distinct mechanism for stabilizing sequential learning under federated constraints, yet each also exposes unresolved challenges such as computational overhead, limited clinical interpretability, and brittleness under highly heterogeneous medical data. Overall, the literature indicates that effective healthcare FCL cannot rely on any single technique in isolation, but instead requires the careful orchestration of architectural, algorithmic, and representational components that jointly respect privacy, heterogeneity, and the operational demands of clinical deployment.

\subsection{IIoT}

Recent work on FCL for IIoT increasingly treats dynamic operating conditions and system constraints as first-class design objectives. Across digital-twin (DT) fault diagnosis, remaining useful life (RUL) prognostics, and industrial process monitoring, the central challenge extends toward sustaining stable performance under heterogeneous data sources, intermittent sensing, and constrained communication. As a result, IIoT-oriented FCL methods typically integrate (i) mechanisms for knowledge retention and robust long-term adaptation, (ii) resource-aware model modularization, and (iii) reliability- or security-aware collaboration strategies.

A first research stream focuses on DT infrastructures, where long-term deployment amplifies the effects of client churn and evolving operational regimes. FCLA-DT \cite{Xia2024FCLADT} demonstrates that continual fault diagnosis in DT-based IIoT systems can be stabilized through task-adaptive parameter aggregation combined with ECC-based group authentication, thereby mitigating unauthorized participation and improving model consistency across sequential tasks. Extending this paradigm, FCLLM-DT \cite{Xia2025} integrates LLM-guided reasoning with federated continual updates to enhance cross-task knowledge retention and adaptive synchronization among distributed DTs. In this context, continual learning serves as a mechanism for maintaining system reliability under sensor failure and cross-factory variability.

A second direction emphasizes trustworthy prognostics. FedCov \cite{Cai2024data} addresses RUL prediction in heterogeneous industrial time-series by coupling a continuous-to-discrete (CDC) transformation with uncertainty-aware evidential aggregation. By explicitly modeling both aleatoric and epistemic uncertainty during training and aggregation, FedCov improves stability under noisy labels and heterogeneous data distributions. Here, uncertainty is treated as an integral component of the learning objective rather than a post-hoc evaluation measure, which is essential for safety-critical predictive maintenance.

Complementary work prioritizes communication and computation efficiency for resource-constrained edge devices. MeCo \cite{Chen2025know} decomposes the learning process into task-invariant meta-knowledge consolidated at the server and task-specific knowledge retrieved from a knowledge pool, enabling forward transfer without heavy replay buffers on devices. This modular formulation reduces communication overhead while preserving continual adaptation across evolving fault domains. Similarly, heterogeneous knowledge-management frameworks \cite{YANG202416} introduce an Adapter--Retainer decomposition in which only lightweight adapter modules are updated, while a frozen backbone preserves historical knowledge. Combined with feature-aware aggregation, this design significantly reduces FLOPs and energy consumption while maintaining long-term stability.

For highly unstable industrial environments, generative replay and gradient-balancing strategies become essential. TRINA \cite{LI2024111491} employs diffusion-based self-challenge replay to preserve invariant structure across heterogeneous industrial conditions, coupled with gradient-balanced aggregation to mitigate update imbalance across clients. These approaches highlight that, in industrial monitoring and soft sensing, robustness depends not only on the backbone architecture but also on principled replay mechanisms and aggregation policies capable of reconciling heterogeneous updates.

Overall, the emerging IIoT FCL literature converges on three core principles: (1) continual adaptation must explicitly support long-term stability; (2) reliability, through uncertainty modeling or reconstruction-based regularization, should be embedded within the optimization process for safety-critical deployments; and (3) scalable industrial deployment requires modular training and selective synchronization that respect edge resource constraints. Collectively, these studies demonstrate that effective industrial FCL constitutes a system-level integration of secure collaboration, robust representation learning, uncertainty-aware modeling, and computationally efficient adaptation, rather than a purely architectural or algorithmic refinement.

\subsection{Smart Cities}

Recent advances in FCL for smart cities
highlight the need for learning systems that operate reliably under dynamic and large-scale data contexts and strict privacy constraints. Across applications such as intrusion detection, intelligent transportation systems, power-grid monitoring, and edge analytics, these studies show that classical FL must evolve toward adaptive and context-aware frameworks capable of operating under adversarial and resource-constrained conditions.

One major research direction focuses on stream-aware optimization for large-scale network analytics. FedStream~\cite{WangNaiyu2024} models heterogeneous data arrival as an inherent source of client variability and stabilizes training through variance-reduced local updates and aggregation schemes that discount stale asynchronous contributions. Complementary work on adaptive cache-controlled streaming FL~\cite{WangHeqiang2024} demonstrates that maintaining long-term performance under storage-constrained routers and gateways requires coordinated design of caching policies and optimization strategies.

A second line of work addresses secure and mobility-resilient learning in intelligent transportation systems. In highly dynamic VANET environments, Particle Swarm Optimization (PSO)-assisted federated ensemble models~\cite{electronics12040894} fuse heterogeneous local classifiers to enable intrusion detection despite intermittent connectivity and sparse data. Ensemble-based aggregation reduces reliance on parameter alignment while improving robustness across distributed vehicle environments.

Another important direction concerns monitoring of cyber--physical infrastructures. FedARF~\cite{Massaoudi2025} combines ensemble-based adaptation with federated aggregation to support real-time stability assessment in power grids processing high-frequency PMU streams. This demonstrates that adaptive ensemble strategies can maintain reliability under rapidly changing system conditions while preserving data privacy.

Continual intrusion detection for IIoT represents a further application of representation-centric FCL. EvoFedIDS~\cite{ZHANG2024108826} employs contrastive learning and prototype-enhanced replay to accommodate emerging attack classes while preserving knowledge of prior threats. By regulating updates along globally important parameter directions, the method maintains discriminative performance despite highly heterogeneous and evolving threat environments.

Beyond algorithmic adaptation, resource-aware control mechanisms have emerged as critical for large-scale deployments. Budget-aware online FL frameworks~\cite{Jin2021} formulate training over stochastic streams as a constrained optimization problem, jointly determining local update frequency and server selection to minimize latency and cost. Such approaches highlight that sustainable smart-city learning systems must explicitly account for bandwidth variability, resource limitations, and operational constraints.

Overall, FCL in security and smart-city contexts is driven by the need for real-time responsiveness and robustness in adversarial environments. Progress in this domain reflects a shift from static aggregation toward integrated solutions that combine adaptive modeling, privacy preservation, mobility resilience, and resource-aware decision-making, enabling cyber--physical systems to operate reliably under persistent uncertainty.

\subsection{Environment}

Recent advances in FCL for 
environment highlight the need for models that can operate under complex and resource-constrained sensing contexts with limited supervision and evolving label spaces. Applications such as hydrological forecasting, waste classification, industrial-metaverse sensing, and multilabel stream analysis demonstrate that sustainability-oriented systems must combine privacy preservation with adaptive learning mechanisms, multimodal fusion, and long-term knowledge retention.

One research direction focuses on hybrid physics--data-driven modeling for ecological prediction. By integrating physics-informed hydrological priors with federated neural surrogates, the approach in~\cite{app131810203} enables robust streamflow estimation across geographically distributed monitoring stations while mitigating cross-site inconsistencies across diverse environmental conditions.

In recycling automation, continual adaptation is essential because visual characteristics of waste materials change over time due to contamination, deformation, and environmental factors. The federated continual vision framework in~\cite{SHAMI2025114976} incrementally updates plastic classifiers using prototype rehearsal and adaptive regularization, maintaining performance across diverse facilities without sharing raw images.

Multimodal sensing environments introduce additional challenges related to cross-modal alignment and communication efficiency. The method in~\cite{YIN2025107345} combines FSM-based multimodal alignment with the FedMAC optimization strategy to preserve semantic consistency across modalities during continual updates. Similarly, the HFedMS framework~\cite{Zeng2023Hfedms} employs semantic compression and memory retention to maintain long-term contextual knowledge while reducing communication overhead, enabling scalable multimodal learning in industrial-metaverse scenarios.

Another important direction addresses evolving data streams with changing feature and label distributions. Semi-supervised FCL methods such as~\cite{Cobbinah2023} integrate pseudo-labeling with adaptive instance selection to support learning under scarce annotations, while multilabel approaches~\cite{Lamptey2025} model dependencies between labels using micro-cluster replay and graph-based techniques to capture joint evolution in feature and label spaces.

Overall, FCL in environmental and sustainability-oriented domains must cope with long-term variability, heterogeneous sensing, and incomplete supervision. Effective solutions typically combine physical priors, semantic memory mechanisms, adaptive multimodal fusion, and efficient communication strategies to support reliable learning over extended time horizons.

Across the surveyed domains, FCL proves most valuable in real-world deployments characterized by complex data environments and strong privacy or resource constraints. Despite domain-specific differences, a common set of practical requirements emerges, including maintaining long-term model reliability, adapting to evolving operational conditions, enabling secure collaboration, and ensuring efficiency on resource-constrained edge devices. These observations highlight the need for hybrid FCL solutions that integrate complementary strategies such as regularization, replay, hierarchical modeling, personalization, multimodal fusion, and uncertainty-aware learning within unified system designs.

\section{Datasets, Evaluation Protocols, and Metrics in FCL}
\label{sec:dataset}

Datasets play a central role in FCL because challenges such as non-IID distributions, temporal drift, and knowledge retention are inherently data-dependent. This section reviews three closely related aspects: \textit{data modalities}, \textit{evaluation protocols}, and \textit{evaluation metrics}. Data modalities determine the structure and dynamics of the input space, while evaluation protocols and metrics define how FCL systems are assessed under continual, distributed, and heterogeneous conditions.

\subsection{Data Modalities in FCL}

Different data modalities in FCL introduce distinct forms of heterogeneity, temporal variation, and communication burden. The most common modalities include image, time-series, text, audio, network traffic, and multimodal data, each of which shapes the learning and evaluation challenges in different ways.

\subsubsection{Image-Based Data}

Image-based datasets are among the most widely used modalities in FCL. A large portion of the literature relies on standard visual benchmarks such as MNIST, Fashion-MNIST, CIFAR-10/100, and TinyImageNet to simulate distributed, sequential, and non-IID settings~\cite{ZHANG2024108826, Cai2024Federated,WU2024414,  LiuWeijie2024,Zhao2024FINAL,Yu2025,WANG2023551, PanWang2025FCCL}.

Beyond synthetic benchmarks, many studies also employ real-world image datasets from domains such as healthcare and industry. Medical imaging datasets, including COVID-XRAY, COVID-CT, OCT, MedPix, and NIH Chest X-ray, are frequently used to model multi-institutional variability, class imbalance, and evolving data distributions~\cite{zhou2025cross, IQBAL2025104157,Sun2025Federated,CFLViT2025,Salman2026}. Other application-specific visual datasets, such as those collected from recycling systems, manufacturing environments, or edge devices, further reflect the practical diversity of image-based FCL settings~\cite{SHAMI2025114976}.


\subsubsection{Sensor and Time-Series Data}

Sensor and time-series data constitute another major modality in FCL, particularly because they naturally reflect temporal evolution, non-stationarity, and concept drift. These datasets often consist of sequential measurements collected from continuously operating systems and are therefore well aligned with federated continual and streaming settings.

Representative examples include physiological signals such as ECG, EEG, PPG, electrodermal activity (EDA), and respiration~\cite{Nandi2023DockerFederated,NANDI2022340,XIONG2026129882,SUN2023586}, as well as industrial and IIoT measurements such as vibration, temperature, pressure, and acoustic signals from machinery and production systems~\cite{WU2024414, LI2024111491,Chen2025know,Xia2024FCLADT,Cai2024data, FCLLM_DT}. Time-series data are also common in energy systems, environmental monitoring, mobile sensing, and smart grids~\cite{Jin2021,casado2023ensemble,Casado2022ConceptDrift,Elgalhud2025, Massaoudi2025,app131810203}. In some cases, network traffic records are similarly treated as sequential streams in distributed environments~\cite{Mawuli2023,WangNaiyu2024}.


\subsubsection{Text and NLP Data}

Textual datasets appear less frequently in FCL studies compared to image and time-series data, but they introduce distinct characteristics related to language variability and semantic structure. These datasets often consist of user-generated or domain-specific text, leading to strong heterogeneity across clients.

Representative datasets include Shakespeare and other text corpora used for language modeling and classification tasks~\cite{electronics14152945,Zheng2025Meta}. These datasets are typically partitioned by users or sources, resulting in personalized and highly non-IID data distributions.
Additional text datasets such as THUCNews are used for large-scale text classification and domain-specific language analysis~\cite{Zhong2025}. In many cases, textual data are also incorporated into heterogeneous or cross-domain datasets, where they coexist with other modalities such as images or graphs~\cite{Luopan2025Loci,YIN2025107345}.

Overall, NLP datasets in FCL are characterized by semantic variability, vocabulary differences, and strong client-level personalization.


\subsubsection{Audio and Speech Data}

Audio and speech datasets form a relatively small subset of FCL data modalities. These datasets consist of speech recordings collected from multiple speakers under diverse acoustic conditions.
Commonly used datasets include VoxCeleb and CnCeleb, which provide large-scale multilingual and multi-speaker audio data~\cite{chen2023learning}. These datasets exhibit variability in speaker identity, recording environment, and language, leading to heterogeneous data distributions across clients.

Speech datasets are inherently sequential and often contain temporal and contextual dependencies. They are typically organized by speaker or domain, making them suitable for distributed settings where each client corresponds to a specific data source.


\subsubsection{Network Traffic and Cybersecurity Data}

Network traffic datasets represent an important modality in FCL, particularly in cybersecurity applications. These datasets consist of packet-level or flow-level records capturing communication patterns and system behavior.

Widely used datasets include CICIDS2017, CICIDS2018, UNSW-NB15, Bot-IoT, and ISCX VPN-nonVPN~\cite{ZHANG2024108826,jin2024fliids}. These datasets contain various types of normal and malicious traffic, often with significant class imbalance and evolving distributions.

In addition, several studies employ streaming network datasets or real-time traffic traces to represent dynamic environments such as IoT networks, vehicular systems, and mobile infrastructures~\cite{WangNaiyu2024,Augello2025,Gelenbe2024,electronics12040894}. These datasets reflect continuously changing traffic patterns and the emergence of new behaviors over time.

Overall, cybersecurity datasets in FCL are characterized by high dimensionality, temporal evolution, and strong heterogeneity across sources.


\subsubsection{Multimodal Data}

Multimodal datasets combine multiple data types, such as image, text, audio, and sensor data, into a unified representation, enabling richer modeling of complex real-world environments.

Representative examples include datasets that integrate physiological signals with contextual information~\cite{Nandi2023DockerFederated,SUN2023586}, as well as heterogeneous datasets combining vision, language, and graph data~\cite{Luopan2025Loci}. Other studies adopt multimodal settings involving image, text, and sensor data across distributed clients~\cite{YIN2025107345}.
A key characteristic of these datasets is heterogeneous feature spaces, modality-specific distributions, and varying data availability across clients. In FCL settings, they further introduce challenges such as modality imbalance, missing modalities, and cross-modal alignment under distributed and non-IID conditions.

\begin{figure}[t]
    \centering
    \includegraphics[width=0.85\linewidth]{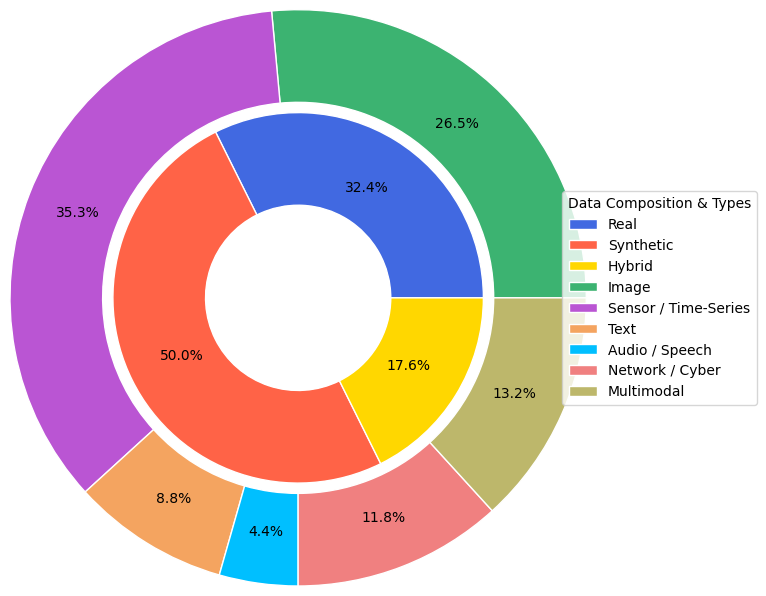}
    \caption{Donut chart summarizing dataset realism (inner ring) and modality categories (outer ring) across all surveyed FCL studies.}
    \label{fig:FCL-dataset-donut}
\end{figure}

To complement the qualitative discussion, Figure~\ref{fig:FCL-dataset-donut} provides a quantitative overview of datasets used in the surveyed FCL literature. The inner ring categorizes dataset types into real, synthetic, and hybrid. Here, ``real'' refers to datasets originating from real-world sources, although in most cases the federated setting itself is simulated by partitioning data across clients. The results indicate a notable prevalence of synthetic and mixed datasets, reflecting the continued reliance on controlled experimental environments.

The outer ring summarizes data modalities, showing that sensor and time-series data are the most dominant, followed by image-based datasets. Other modalities, including text, audio, cybersecurity (network traffic), and multimodal data, appear less frequently but demonstrate a growing trend in recent FCL research.

\subsection{Evaluation Protocols in FCL}

Evaluation protocols in FCL define how models are assessed under continual and distributed settings, including evaluation strategies and federated configurations. These protocols determine how performance is measured over time and across heterogeneous clients.

In practice, FCL evaluation protocols can be organized along two complementary dimensions: (i) how performance is measured over time, and (ii) the federated conditions under which evaluation is conducted.

\subsubsection{Evaluation Strategies}

These evaluation strategies are typically applied across different CL paradigms discussed in Section~\ref{sec:taxonomy}.

\paragraph{Per-Task Evaluation}
In per-task evaluation, model performance is measured after completing each task, allowing a clear assessment of knowledge retention across sequential tasks. This strategy is commonly used in task- and class-incremental settings, particularly in studies based on benchmark datasets such as CIFAR and MNIST~\cite{ZHANG2024108826,Zhao2024FINAL,WANG2023551,Yu2025,PanWang2025FCCL,SHAMI2025114976,Salman2026}.

\paragraph{Per-Round Evaluation}
Per-round evaluation measures model performance at each communication round during federated training. It is widely adopted in streaming and incremental settings to monitor training dynamics and convergence behavior over time~\cite{Cai2024Federated,WU2024414,WangHeqiang2024,BACCARELLI2022376,Su2024}.

\paragraph{Test-then-Train (Online Evaluation)}
In test-then-train evaluation, the model is first evaluated on incoming data and then updated using the same data, reflecting realistic online learning conditions. This strategy is mainly used in streaming scenarios with sequential data arrival, such as physiological and sensor data streams~\cite{Nandi2023DockerFederated,NANDI2022340,Cobbinah2023}.

\paragraph{Continuous / Time-based Evaluation}
Continuous evaluation tracks model performance over time as new data arrive, making it suitable for streaming and non-stationary environments. It is frequently applied in time-series and real-time systems such as IoT, energy, and network data streams~\cite{Jin2021,casado2023ensemble,WangNaiyu2024,XIONG2026129882,Casado2022ConceptDrift,Massaoudi2025}.

\paragraph{Multi-Scenario Evaluation}
Multi-scenario evaluation examines model performance across multiple continual learning settings, such as task-, domain-, and time-evolving scenarios. This approach is typically used in complex and hybrid environments involving heterogeneous datasets~\cite{Nandi2023DockerFederated,NANDI2022340,Zhong2025}.

\subsubsection{Federated Evaluation Settings}

\paragraph{Global vs Local Evaluation}
Global evaluation measures the performance of the aggregated model on a shared test set, providing an overall system-level perspective. In contrast, local evaluation assesses performance on each client’s private data, capturing client-specific behavior and data heterogeneity~\cite{ZHANG2024108826,Cai2024Federated,WangHeqiang2024,BACCARELLI2022376,PanWang2025FCCL}.

\paragraph{Centralized vs Decentralized Evaluation}
In centralized evaluation, model assessment is performed using a common validation or test dataset, typically maintained at the server side. Decentralized evaluation relies on client-side testing, where each client evaluates the model locally without sharing raw data, aligning with privacy-preserving constraints~\cite{Nandi2023DockerFederated,NANDI2022340,WU2024414,WangNaiyu2024}.

\paragraph{Synchronous vs Asynchronous Settings}
Synchronous settings evaluate models when all participating clients update simultaneously in each round. In contrast, asynchronous settings allow clients to update and communicate at different times, reflecting more realistic conditions with varying availability and communication delays~\cite{WU2024414,Chen2025know,FCLLM_DT}.

\paragraph{Client Participation and Heterogeneity}
Evaluation settings also consider client participation and heterogeneity, where only a subset of clients may participate in each round and data distributions differ significantly across clients. These conditions are commonly modeled in FCL studies to reflect non-IID and dynamic real-world environments~\cite{Nandi2023DockerFederated,Jin2021,XIONG2026129882,Elgalhud2025,Massaoudi2025}.

\subsection{Evaluation Metrics}

Evaluation metrics in FCL quantify model performance from multiple complementary perspectives, including predictive accuracy, knowledge retention, system efficiency, and robustness. Since FCL combines continual and federated learning, no single metric is sufficient; instead, multiple metrics are typically reported to capture different aspects of model behavior under non-stationary and distributed conditions.

\paragraph{Performance Metrics}
Standard classification metrics such as accuracy, F1-score, precision, recall, and AUC are commonly used to evaluate predictive performance across tasks, classes, and clients. In imbalanced scenarios, such as medical diagnosis or intrusion detection, F1-score and recall often provide more informative insights than accuracy alone.

\paragraph{CL Metrics}
CL metrics evaluate how well the model preserves and transfers knowledge over time. Common measures include forgetting rate, backward transfer (BWT), and forward transfer (FWT), which respectively quantify performance degradation on previously learned tasks, the influence of new learning on past knowledge, and the benefit of prior experience for future tasks.

\paragraph{FL Metrics}
FL metrics focus on system-level efficiency and scalability, including communication cost, convergence speed, latency, computation time, and memory overhead. These metrics are particularly important in resource-constrained and partially participating federated environments.

\paragraph{Reliability and Domain-Specific Metrics}
Additional metrics are used to assess robustness and application-specific requirements, including RMSE/MSE for regression tasks, uncertainty estimation, fairness across clients, calibration, and stability under distribution drift. These measures are especially important in safety-critical domains, where reliability and trustworthiness are essential.

Table~\ref{tab:fcl_summary_1} summarizes representative FCL studies by jointly comparing application domains, data modalities, CL paradigms, evaluation strategies, and metrics. This unified view highlights the diversity of real-world scenarios and the heterogeneity of evaluation practices across domains such as healthcare, IoT, and cybersecurity.

For clarity, we adopt compact notations for both learning settings and metrics. Specifically, TI, CI, DI, SD, HY, and P denote task-incremental, class-incremental, domain-incremental, streaming, hybrid, and personalized settings, respectively. Evaluation metrics are reported in abbreviated form, including Acc (accuracy), F1 (F1-score), Prec. (precision), Rec. (recall), and AUC (Area Under Curve), as well as RMSE (Root Mean Square Error), MSE (Mean Square Error), and MAE (Mean Absolute Error) for regression tasks. CL behavior is captured through metrics such as Forget (forgetting rate), Backward Transfer (BWT) and Forward Transfer (FWT), while federated efficiency is reflected by Comm. (communication cost), Conv. (convergence speed), Lat. (latency), and Time (overhead).

Overall, the table shows that most studies emphasize accuracy-based evaluation, while fewer works explicitly consider CL metrics. In contrast, communication cost and convergence are commonly reported, underscoring the importance of system-level efficiency in federated settings. These observations indicate the need for more comprehensive and standardized evaluation practices in FCL.

\begin{landscape}
\begin{table}[t]
\centering
\scriptsize
\setlength{\tabcolsep}{3pt}
\renewcommand{\arraystretch}{1.5}
\caption{Representative FCL studies across application domains, use case / task, CL paradigms, data modalities, dataset type, evaluation settings, core mechanisms, strengths and limitations.}
\label{tab:fcl_summary_1}
\fontsize{10}{10}\selectfont

\resizebox{\linewidth}{!}{
\begin{tabular}{llllllllllllll}
\toprule
\textbf{Ref.} & \textbf{Application Domain} & \textbf{Use Case / Task} & \textbf{CL} & \textbf{Data Modality} & \textbf{Dataset Type} & \textbf{Eval. Strategy} & \textbf{Fed.} & \textbf{Evaluation Metrics} & \textbf{Core Mechanism} & \textbf{Strength} & \textbf{Limitation} \\
\midrule

\cite{PanWang2025FCCL} & Healthcare & Image classification & CI & Image & Sim & Per-task & Global & Acc; Micro-F1 & Adapters + distillation & Low forgetting & Storage overhead \\

\cite{ID1,jin2024fliids} & Cybersecurity & Intrusion detection & CI & Time-series & Sim & Per-task & Global & Acc; F1; Prec.; Rec.; Det. Rate & Memory + bias correction & Handles new attacks & Scalability \\

\cite{Sun2025Federated,Iqbal2026} & Healthcare & Disease diagnosis & CI & Image+TS & Mixed & Per-task & Global & Acc; RMSE; MSE; Forget.; FID; FWT & Diffusion replay & Robust to non-IID & High cost \\

\cite{Zheng2025Meta,Zhong2025} & Edge AI & Personalized classification & CI+HY & Multi & Mixed & Continuous & Global+local & Acc; Forget.; Comm.; Lat.; Storage & Meta + replay & Personalization & Overhead \\

\cite{chen2023learning,IQBAL2025107920,zhang2025fedstil} & Vision/audio & Re-ID / speech & DI+TI & Multi & Mixed & Per-task & Global & Rank-1 Acc; mAP; AUC; F1; Comm. & Distillation + transfer & Domain robustness & Complexity \\

\cite{ZHANG2024108826,Zhao2024FINAL,Yu2025,WANG2023551} & IoT & Sequential classification & TI & Image & Sim & Per-task & Global & Acc; Avg. Acc; Forget.; Comm.; Conv. & Fisher + regularization & Efficient comm. & Extra cost \\

\cite{zhou2025cross,IQBAL2025104157,CFLViT2025} & Healthcare & Medical diagnosis & TI+DI & Image & Mixed & Per-task & Global & Acc; F1; Forget.; Fair.; Conv. & EWC + distillation & Stable adaptation & Heavy compute \\

\cite{Li2023per,electronics14152945} & Biometrics & Personalized ID & TI+P & Multi & Mixed & Per-task & Global+local & Acc; Forget.; BWT; RA; Comm. & Meta-learning & Personalization & High cost \\

\cite{YANG202416,SHAMI2025114976} & Recycling & Plastic sorting & TI & Image & Mixed & Per-task & Global & Acc; Forget.; FLOPs; Comm.; Energy & Adapter+replay & Low FLOPs & Tuning needed \\

\cite{LI2024111491,Liu2025Spars,Chen2025know,Luopan2025Loci} & Industry & Fault diagnosis & TI+DI & Multi & Real & Continuous & Global & Acc; RMSE; MAE; Forget.; Comm.; Memory & Meta + replay & Handles heterogeneity & Complexity \\

\cite{Nandi2023DockerFederated,LiuWeijie2024,YIN2025107345} & IoMT & Multimodal learning & HY & Multi & Mixed & Multi-scenario & Global & Acc; F1; Conv.; Comm. & Meta + attention & Robust & Overhead \\

\cite{Nandi2023DockerFederated,NANDI2022340} & Healthcare / IoMT & Emotion recognition & SD & Time-series & Sim & Online & Global & Acc & Online + fusion & Real-time adaptation & Limited interpretability \\

\cite{Cai2024Federated,Cai2024Boosting} & Privacy FL & Data release / classification & SD & Multi & Sim & Per-round & Global & Acc; RMSE; MSE; Comm.; Time & DP aggregation & Privacy-utility trade-off & Complexity \\

\cite{XIONG2026129882,Salman2026} & Medical IoT & Online diagnosis & SD & Image+TS & Mixed & Continuous & Global & Acc; Prec.; Rec.; F1; Comm.; Robust. & Bayesian / distillation & Robust + efficient & Overhead \\

\cite{Jin2021,BACCARELLI2022376,Su2024} & IoT & Control / scheduling & SD & Time-series & Real & Continuous & Global & Conv.; Lat.; Comm.; Utility & Async + Lyapunov & Low latency & Optimization complexity \\

\cite{WU2024414,Zeng2023Hfedms,Xia2024FCLADT,FCLLM_DT} & IIoT & Fault diagnosis & SD & Multi & Real & Continuous & Global & Acc; F1; Comm.; Conv.; Train Eff. & Replay + semantic comp. & Industrial robustness & High compute \\

\cite{casado2023ensemble,Casado2022ConceptDrift,Massaoudi2025} & Smart grid & Drift detection & SD & Time-series & Real & Continuous & Global & Acc; Bal. Acc; Sens.; Spec.; Comm. & Drift + ensemble & Handles drift & Ensemble cost \\

\cite{ZHANG2024108826,Mawuli2023,WangHeqiang2024,WangNaiyu2024,Augello2025,Gelenbe2024,electronics12040894} & Cybersecurity & Intrusion / traffic detection & SD & Network & Mixed & Continuous & Global & Acc; F1; Prec.; Rec.; Comm.; Conv. & Replay + prototype & Robust to evolving threats & Storage overhead \\

\cite{Le2021,Cobbinah2023,Lamptey2025,fini2022self,padilla2025tefes} & Edge AI & Representation learning & SD & Multi & Mixed & Continuous & Global & Acc; F1; Prec.; Rec.; Comm.; Reliab. & Self-supervised replay & Works with unlabeled data & Limited interpretability \\

\cite{Elgalhud2025,app131810203} & Energy & Forecasting & SD & Time-series & Real & Continuous & Global & RMSE; MAE; Forecast Acc; Train Time & Sequential / hybrid & Adaptive prediction & Comm. overhead \\

\cite{HU2025,Zhu2023,Tian2026,Niu2025DPFL,dalfabbro2024shed} & Secure FL & Optimization / privacy & SD & Multi & Mixed & Continuous & Global & Acc; F1; Privacy; Conv.; Comm. & DP + optimization & Strong guarantees & High cost \\
\bottomrule
\end{tabular}
}

\vspace{2mm}

\end{table}
\end{landscape}

\section{Limitations, Open Challenges, and Future Directions}
\label{sec:limitations-FCL}

Despite the rapid progress of FCL, the surveyed literature reveals several fundamental limitations that restrict the maturity and real-world applicability of current approaches. These limitations are not isolated technical issues but rather structural gaps that span benchmarking, robustness, personalization, trustworthiness, and system efficiency.

A persistent weakness of existing studies is the lack of realistic and unified benchmarking protocols. Most works rely on simulated federated environments constructed by partitioning centralized datasets, which only partially capture real-world characteristics such as asynchronous participation, intermittent connectivity, device failures, and long-term data evolution~\cite{Nandi2023DockerFederated,Towards2024Towards,Casado2022ConceptDrift}. Even among studies that explicitly address streaming data, experimental protocols differ widely in their definitions of tasks, temporal windows, and evaluation schedules, which severely limits cross-paper comparability~\cite{WangHeqiang2024, Casado2022ConceptDrift}. Several frameworks demonstrate strong performance on specific datasets or domains, such as physiological emotion streams~\cite{Nandi2023DockerFederated,NANDI2022340}, industrial image benchmarks~\cite{ID6}, or controlled IoT environments~\cite{Towards2024Towards}, but their conclusions remain difficult to generalize. This fragmentation highlights the need for standardized FCL benchmarks that simultaneously model temporal evolution, client heterogeneity, communication dynamics, and system constraints. Without such benchmarks, progress in long-term memory retention~\cite{PanWang2025FCCL} and heterogeneous-task learning~\cite{ID30} will remain difficult to quantify in a comparable manner.
Another critical challenge lies in robust adaptation under concept drift and spatiotemporal heterogeneity. Existing approaches rely on drift detection, rehearsal mechanisms, cache management strategies, prototype maintenance, or diffusion-based replay to cope with non-stationary data~\cite{WangHeqiang2024,Casado2022ConceptDrift}. However, most of these methods characterize drift only at a statistical level and do not explain its underlying semantic causes, which limits interpretability and prevents actionable responses in safety-critical systems~\cite{WangHeqiang2024,Casado2022ConceptDrift}. In practice, drift is often treated as a numerical anomaly rather than as a semantic change in the data-generating process.

Moreover, current drift-handling mechanisms are typically designed for single-modality and fully supervised streams, whereas real-world deployments frequently involve multimodal data, weak supervision, delayed labels, or evolving label spaces~\cite{Cobbinah2023,Lamptey2025}. Bridging this gap requires drift modeling strategies that jointly capture distributional shifts, semantic changes in labels, and uncertainty dynamics, especially when modalities or task definitions evolve over time.

More general FCL frameworks that unify semantic drift modeling, uncertainty-aware adaptation, and multimodal streaming remain largely unexplored. Bayesian and uncertainty-driven approaches provide a promising direction by enabling confidence-calibrated updates, risk-aware adaptation, and safer continual learning under non-stationary and heterogeneous environments~\cite{XIONG2026129882,Cai2024data}.

Scalable personalization under extreme client heterogeneity remains a major open challenge in FCL. 
Existing studies introduce personalization through meta-learning, adapter modules, parameter decomposition, or client-specific classifiers~\cite{PanWang2025FCCL}. 
Although these mechanisms significantly improve local adaptability and mitigate catastrophic forgetting, they often come at the cost of increased computational, memory, and communication overhead. 
For instance, adapter-based methods may accumulate substantial memory footprints as the number of tasks grows~\cite{PanWang2025FCCL}, while meta-learning and knowledge-pool based approaches introduce considerable system complexity and training instability~\cite{Zheng2025Meta}. 

In contrast, fully shared global models are computationally simpler and easier to deploy, but their performance rapidly degrades under strong statistical heterogeneity and personalized data distributions~\cite{Towards2024Towards}. 
This trade-off reveals a fundamental tension between personalization and scalability in FCL systems.
Addressing this tension requires an adaptive mechanism that dynamically regulates the degree of personalization according to task similarity, client diversity, and system resources constraints.

Future FCL frameworks must therefore become inherently resource-aware, dynamically balancing global knowledge sharing and local specialization according to device capabilities, task similarity, and the severity of distributional drift. 
Lightweight modular architectures that decouple stable shared representations from small adaptive components, combined with selective synchronization and task-aware routing strategies, offer a principled pathway toward scalable personalization while preserving communication efficiency and long-term stability.

Trustworthiness under privacy and security constraints constitutes another critical and still unresolved dimension of FCL. 
While techniques such as differential privacy and continual data-release mechanisms significantly strengthen confidentiality guarantees, they inevitably introduce privacy--utility trade-offs and additional computational overhead, which become increasingly severe over long training horizons~\cite{HU2025}. 
Moreover, many existing FCL systems still operate without any explicit privacy protection, even when applied to highly sensitive domains such as healthcare, industrial monitoring, or surveillance~\cite{Zhao2024FINAL}. 

From a security perspective, threats including data poisoning, backdoor attacks, and adversarial manipulation of continual updates remain only partially explored, and most defenses are designed for narrow, domain-specific scenarios rather than general-purpose FCL deployments~\cite{ZHANG2024108826}. 
As a result, current evaluations rarely consider privacy, robustness, and forgetting mitigation in a unified manner, leaving a significant gap between algorithmic design and real-world trust requirements.

Future FCL frameworks must therefore integrate privacy preservation, security robustness, and continual knowledge retention into a single principled evaluation and design paradigm. 
Promising directions include privacy-preserving replay buffers, differentially private synthetic data generation for continual rehearsal, and explanation-level integrity verification to detect malicious or corrupted updates. 
Such mechanisms would enable FCL systems to become not only adaptive and scalable, but also trustworthy by design, integrating privacy, robustness, and continual retention within a unified framework.

Efficiency and deployability on resource-constrained edge devices constitute another major bottleneck for the practical scalability of FCL systems. 
Many continual learning mechanisms, including replay buffers, prototype memories, Fisher information accumulation, diffusion-based generators, and Bayesian aggregation, introduce substantial computational and memory overhead, which limits their feasibility on lightweight clients~\cite{Zhao2024FINAL}. 
While communication-efficient strategies such as bidirectional compression~\cite{Zhang2023FedSI}, adaptive aggregation scheduling~\cite{LiuWeijie2024}, sparse topology learning~\cite{Liu2025Spars}, and budget-aware optimization~\cite{Jin2021} partially reduce system costs, they are rarely investigated together with continual memory and forgetting-mitigation mechanisms.

Moreover, most FCL studies remain evaluated in simulated environments and overlook realistic deployment constraints such as device heterogeneity, energy consumption, storage limitations, and streaming execution frameworks. 
Only a limited number of works consider end-to-end implementations using practical infrastructures, for instance through Docker-based federated platforms or streaming orchestration systems~\cite{Nandi2023DockerFederated,Towards2024Towards}. 
This gap highlights that many current FCL solutions remain algorithmically sound but systemically impractical.

Future research must therefore move toward holistic system-level design, where memory, computation, and communication are co-optimized under continual learning objectives. 
Promising directions include modular architectures that decouple stable, globally shared representations from lightweight and adaptive local components, enabling efficient personalization and retention without excessive resource consumption~\cite{PanWang2025FCCL}. 
Such designs offer a principled trade-off between efficiency, scalability, and long-term knowledge preservation, which is essential for deploying FCL in real-world edge environments.

Learning under limited, evolving, or weak supervision constitutes another critical gap in current FCL research. 
Most existing approaches implicitly assume fully labeled and reliably annotated data streams, whereas real-world federated environments are often characterized by partial labels, delayed annotations, or entirely unlabeled data. 
This mismatch significantly limits the applicability of many FCL methods in practical deployments, especially in safety-critical and large-scale monitoring scenarios.

Recent semi-supervised and self-supervised approaches provide promising initial steps by enabling continual adaptation through prototype consistency learning, pseudo-labeling, or representation learning from unlabeled streams~\cite{Cobbinah2023}. 
However, these methods remain relatively underexplored in federated continual settings and are often evaluated under simplified assumptions.

The challenge becomes even more complex in multilabel and evolving-label scenarios, where label dependencies, correlations, and class semantics change over time~\cite{Lamptey2025}. 
In such cases, catastrophic forgetting is not limited to feature representations but also affects the structure of the label space itself, making long-term stability significantly harder to maintain.

These observations suggest that future FCL frameworks must natively support weak supervision, delayed feedback, and open-set class emergence. 
Such capabilities are particularly crucial in domains such as security monitoring, healthcare diagnostics, and industrial fault detection, where novel events, rare conditions, and incomplete annotations are the norm rather than the exception~\cite{ZHANG2024108826,Lamptey2025}. 
Developing robust mechanisms that jointly address label uncertainty, representation stability, and continual adaptation therefore represents a key direction for advancing FCL toward real-world usability.

Overall, the limitations highlighted in the existing literature suggest that FCL must move beyond isolated algorithmic innovations toward a holistic, system-level research paradigm. 
Future progress will critically depend on the establishment of standardized benchmarks and principled evaluation protocols that simultaneously account for catastrophic forgetting, robustness to concept drift, privacy and security guarantees, and system efficiency. 
Moreover, FCL architectures must be designed to balance adaptability, trustworthiness, and deployability, rather than optimizing each aspect in isolation~\cite{WangHeqiang2024, HU2025,Casado2022ConceptDrift}. 
Only through such integrated design and evaluation frameworks can FCL mature into a practical foundation for large-scale, real-world decentralized continual learning systems.

Recent advances in Large Language Models (LLMs) introduce a promising direction for FCL. LLMs provide strong capabilities in representation learning, knowledge transfer, and generative modeling, which can potentially mitigate several core challenges in FCL, including catastrophic forgetting, data scarcity, and heterogeneous data distributions.
Recent studies have started to explore the integration of LLMs within FCL systems, particularly in data-driven and industrial settings. For instance, LLMs combined with retrieval-augmented generation (RAG) have been used to generate synthetic data streams and compensate for missing or corrupted sensor data in Industrial IoT environments, enabling more stable continual learning under non-stationary conditions~\cite{FCLLM_DT}. These approaches demonstrate that LLMs can act as auxiliary knowledge generators, supporting continual adaptation when real data is incomplete, delayed, or unreliable.
Moreover, LLMs can facilitate semantic-level knowledge sharing across clients, enabling more effective cross-client generalization and alignment in highly heterogeneous federated environments. Their ability to capture high-level abstractions makes them particularly suitable for multimodal and weakly supervised scenarios, where traditional FCL methods struggle.
However, integrating LLMs into FCL introduces significant challenges. 
The large scale of LLMs leads to substantial computational and communication overhead, which is incompatible with resource-constrained edge devices. In addition, continual fine-tuning of LLMs in decentralized settings raises concerns regarding stability, privacy leakage, and model drift over time. Efficient adaptation strategies, such as parameter-efficient fine-tuning, modular architectures, and selective knowledge distillation, are therefore essential.
Overall, the integration of LLMs into FCL represents a highly promising yet still underexplored research direction. Developing scalable, privacy-preserving, and resource-efficient mechanisms for incorporating LLMs into FCL systems remains an important avenue for future work.

\section{Conclusion}
\label{sec:conclusion}

This survey has presented a comprehensive overview of Federated Continual Learning (FCL) as an emerging paradigm that integrates the strengths of Federated Learning (FL) and Continual Learning (CL) to support adaptive and privacy-preserving learning over distributed, non-stationary data. We showed that the stationarity assumption underlying classical FL rarely holds in real-world deployments, while centralized CL methods fail to address decentralization, privacy constraints, and system heterogeneity. FCL naturally bridges these gaps by enabling long-term knowledge retention and continual adaptation across distributed clients.

We provided a unified problem formulation and a multi-dimensional taxonomy of FCL methods, highlighting key design dimensions including adaptation strategies, aggregation mechanisms, personalization techniques, and CL paradigms. Our analysis revealed that no single approach can address all aspects of FCL, and that effective solutions typically combine complementary mechanisms such as regularization, replay, modular architectures, sparse adaptation, and meta-learning.

A wide range of application domains, including healthcare, IIoT, cybersecurity, networking, smart cities, and environmental monitoring, demonstrates the strong practical importance of FCL. These scenarios emphasize the need for evaluation frameworks that capture not only accuracy, but also long-term stability, resistance to forgetting, robustness to distributional drift, communication efficiency, and memory usage.

Despite significant progress, major challenges remain, including extreme heterogeneity under temporal drift, privacy-preserving memory design, long-term model stability, fairness across clients and time, and the lack of standardized benchmarks. Addressing these issues will require deeper integration between learning theory, distributed systems, privacy-preserving techniques, and resource-aware optimization.

In summary, FCL represents a critical step toward autonomous and sustainable learning systems capable of operating in dynamic and privacy-sensitive environments. By clarifying the conceptual foundations, organizing existing research, and identifying key opportunities, this survey aims to guide future developments toward robust, fair, and efficient lifelong FL systems.

\section*{Acknowledgements}
This work has been partly funded by the PNRR - M4C2 - Investimento 1.3, Partenariato Esteso PE00000013 - ``FAIR - Future Artificial Intelligence Research'' - Spoke 1 ``Human-centered AI'' under the NextGeneration EU programme, and the Italian Ministry of University and Research (MUR) in the framework of the FoReLab and CrossLab projects (Departments of Excellence).
 
\section*{Data Availability}
No new data were generated or analyzed in this study. All data discussed in this article are derived from previously published studies.



\bibliographystyle{elsarticle-num}
\bibliography{references}

\end{document}